\documentclass[sigconf]{acmart}
\usepackage{multirow}
\usepackage{hyperref}

\usepackage{amsmath}

\usepackage{amssymb}
\usepackage{mathtools}
\usepackage{amsthm}
\usepackage{pifont}
\usepackage{microtype}
\usepackage{graphicx}
\usepackage{bbding} %
\usepackage{booktabs} %
\usepackage{algorithm}
\usepackage{algorithmic}
\usepackage{todonotes}
\usepackage{subcaption}
\usepackage{makecell}
\usepackage{balance}

\AtBeginDocument{%
  }

\setcopyright{acmlicensed}
\copyrightyear{2024}
\acmYear{2024}
\setcopyright{acmlicensed}\acmConference[MM '24]{Proceedings of the 32nd
ACM International Conference on Multimedia}{October 28-November 1,
2024}{Melbourne, VIC, Australia}
\acmBooktitle{Proceedings of the 32nd ACM International Conference on
Multimedia (MM '24), October 28-November 1, 2024, Melbourne, VIC, Australia}
\acmDOI{10.1145/3664647.3680960}
\acmISBN{979-8-4007-0686-8/24/10}

\newcommand{\name}{WeakSAM}
\begin{document}

\title{\name{}: Segment Anything Meets Weakly-supervised Instance-level Recognition}

\author{Lianghui Zhu}
\authornote{Both authors contributed equally to this research.}
\email{lhzh@hust.edu.cn}
\orcid{0009-0009-4899-9320}
\author{Junwei Zhou}
\authornotemark[1]
\email{u202114054@hust.edu.cn}
\affiliation{%
  \institution{School of EIC, Huazhong University of Science \& Technology }
  \city{Wuhan}
  \state{Hubei}
  \country{China}
}

\author{Yan Liu}
\email{goodmandou@gmail.com}
\author{Xin Hao}
\email{chengfeng.hx@antgroup.com}
\affiliation{%
  \institution{Alipay Tian Qian Security Lab}
  \city{Beijing}
  \country{China}}

\author{Wenyu Liu}
\email{liuwy@hust.edu.cn}
\author{Xinggang Wang}
\authornote{Xinggang Wang is the corresponding author.}
\email{xgwang@hust.edu.cn}
\affiliation{%
  \institution{School of EIC, Huazhong University of Science \& Technology}
  \city{Wuhan}
  \state{Hubei}
  \country{China}}

\renewcommand{\shortauthors}{Lianghui Zhu et al.}

\begin{abstract}
Weakly-supervised visual recognition using inexact supervision is a critical yet challenging learning problem. It significantly reduces human labeling costs and traditionally relies on multi-instance learning and pseudo-labeling. This paper introduces WeakSAM and solves the weakly-supervised object detection (WSOD) and segmentation by utilizing the pre-learned world knowledge contained in a vision foundation model, i.e., the Segment Anything Model (SAM).
WeakSAM addresses two critical limitations in traditional WSOD retraining, i.e., pseudo ground truth (PGT) incompleteness and noisy PGT instances, through adaptive PGT generation and Region of Interest (RoI) drop regularization. It also addresses the SAM's shortcomings of requiring human prompts and category unawareness in object detection and segmentation.
Our results indicate that WeakSAM significantly surpasses previous state-of-the-art methods in WSOD and WSIS benchmarks with large margins, i.e. average improvements of 7.4\% and 8.5\%, respectively.
Code and models are released at \url{https://github.com/hustvl/WeakSAM}.
\end{abstract}



\maketitle

\section{Introduction}
\label{sec:intro}

Weakly-supervised learning (WSL)~\cite{zhou2018wsl_intro, wang2013mmmil, xu2014lmws} is a crucial component of machine learning. It is particularly valuable in tasks where strong supervision is difficult to annotate due to the high cost of data labeling~\cite{locatello2020wscomp, schroeter2019wstemp, fu2020wstriplet, duan2022rda, duan2023towards, duan2024mutexmatch, duan2024roll}. 
Due to the massive demand for annotated data in visual perception, WSL is essential in developing a label-efficient recognition system.
In the standard weakly-supervised visual perception paradigm~\cite{tang2018pcl, sui2022sos-wsods, mmChen2021wsol, mmXu2022wsol, mmTan2020wsol, mmShao2021wsol, mmXu2022wsss, mmChen2021wsss, mmQian2022wsss, mmZhang2021wsss, mmZhang2022wsss, mmYang2020wsod}, training commences with inexact supervision, such as image-level labels. 
Subsequently, the trained WSL network is employed to generate pseudo ground truth (PGT), which serves as a form of refined, albeit still inaccurate supervision.
Finally, the PGT is used as inaccurate supervision to launch WSL retraining.
Although the iterative WSL process achieves significant progress, it is still limited by the lack of external knowledge, which restricts the performance of WSL and hinders it from matching fully-supervised learning (FSL).

\begin{figure}[tp]
    \centering
    \includegraphics[width=.85\linewidth]{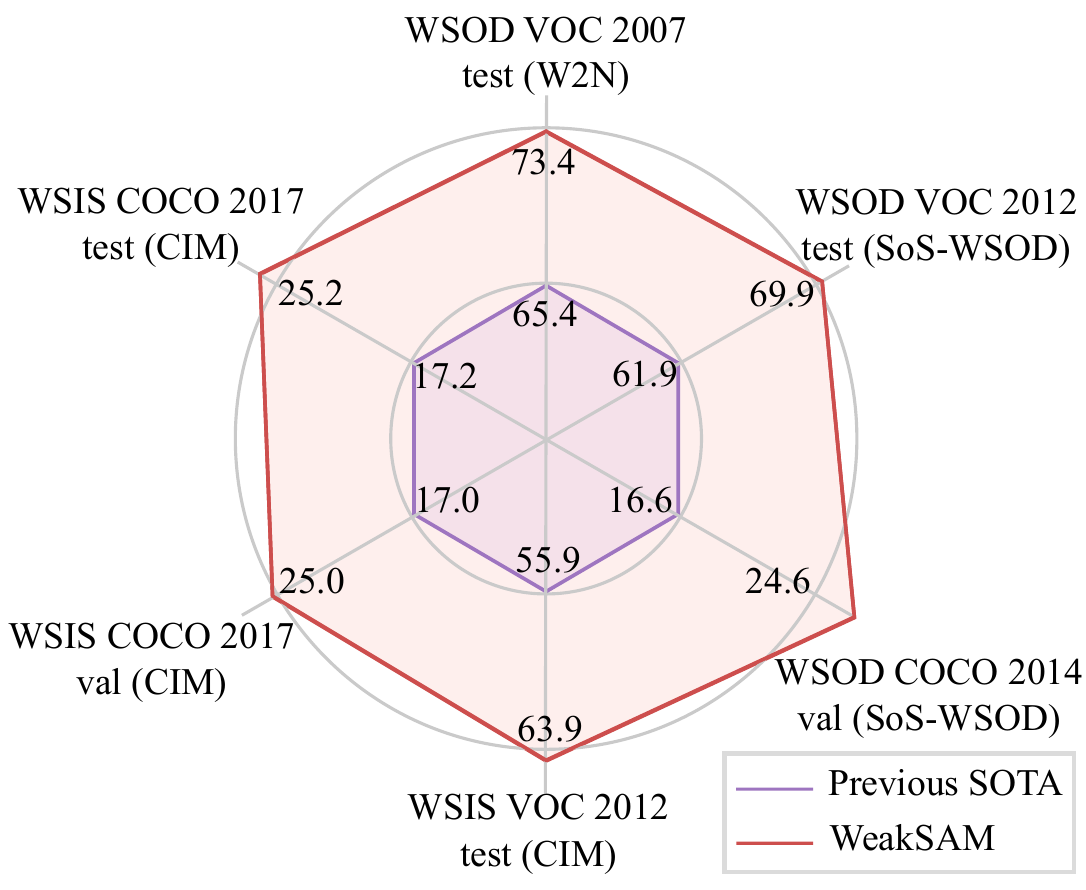}
    \vspace{-0.05 in}
    \caption{
    Quantitative comparisons between \name{} and previous SOTA methods under different tasks and benchmarks. The scale of each axis in the radar chart is normalized by the performance of the previous SOTA methods (marked in parentheses), and the stride of each axis is the same.
    }
    \label{fig:sota_fig}
\vspace{-.2 in}
\end{figure}
Nowadays, foundation models are gaining increasing attention because of their transferable pre-learned world knowledge, which can be regarded as powerful external knowledge for WSL. 
As a vision foundation model, SAM~\cite{kirillov2023sam} achieves outstanding performance in interactive, class-agnostic segmentation. 
SAM owes its success to promptable training on a large-scale dataset. 
However, there are two main drawbacks to SAM: First, SAM requires interactive operations as input, which means it cannot work automatically without human prompts. 
Second, SAM produces class-agnostic segments and cannot assign class labels. 
These drawbacks severely restrict the application of SAM as a direct and generic visual framework.
As a strong complement, WSL is good at mining classification clues through inexact supervision, which can provide automatic prompts for SAM. Subsequently, WSL with SAM's knowledge can further bring class-aware perception.

This motivates us to assimilate SAM within the WSL paradigm. The WeakSAM framework is designed to harness transferable knowledge from SAM, thereby enriching the WSL process. Simultaneously, it offers the capability to deliver automatic classification clues to SAM. This bidirectional enhancement constructs a promising foundation-model-based weakly-supervised visual perception framework. Specifically, in a weakly-supervised object detection (WSOD) setting, WeakSAM uses classification clues as SAM prompts to produce proposals automatically. These proposals are then used in WSOD training for class-aware perception.

Within the scope of the WeakSAM framework, our analysis identifies two prevailing limitations in the iterative WSOD retraining approach: the issue of pseudo ground truth (PGT) incompleteness and the presence of noisy PGT instances. The former, PGT incompleteness, refers to the tendency of WSOD-generated PGT to omit some objects or categories, leading to insufficient training for these categories. The latter, noisy PGT instances, pertain to the prevalent presence of noise within the PGT, which adversely impacts the retraining process. To effectively mitigate these challenges, we introduce two key strategies: adaptive PGT generation to address the PGT incompleteness problem, and Region of Interest (RoI) drop regularization to counteract the noise in PGT instances. Moreover, WeakSAM's capability enables the extension in the realm of weakly-supervised instance segmentation (WSIS). In this context, SAM is employed to further refine WeakSAM-PGT, enabling the generation of pseudo instance segmentation labels. This approach exemplifies WeakSAM is promising to build a unified weakly-supervised instance-level recognition framework.

The main contributions of this paper are as follows:

\begin{itemize}
\item We propose a weakly-supervised instance-level recognition framework (\name{}), which automatically prompts SAM by classification clues for proposals. 
The \name{} proposals reduce the generation time by 65.5\% and improve the recall (IoU=0.9) by 22.9\%, compared to Selective Search~\cite{Uijlings2013SelectiveSF}.
\item We analyze the weaknesses in traditional WSOD retraining, and propose adaptive PGT generation and RoI drop regularization to address them, respectively. After the WeakSAM-WSOD is complete, the proposed WeakSAM can be easily applied to WSIS further.
\item The proposed \name{} achieves state-of-the-art (SOTA) results on the WSOD and WSIS benchmarks, significantly surpassing previous SOTA methods as shown in Fig.~\ref{fig:sota_fig}.
\end{itemize}

\section{Related Work}
\label{sec:relatedw}
\subsection{Segment Anything Model}
The Segment Anything Model (SAM)~\cite{kirillov2023sam} is trained on SA-1B with over 1 billion masks, following the model-in-the-loop manner. Besides, SAM is applied in many visual tasks, e.g., FGVP~\cite{yang2023fgvp} incorporates SAM to achieve zero-shot fine-grained visual prompting, MedSAM~\cite{ma2023medsam} adapts SAM into a large scale medical dataset to build a medical foundation model, and some methods~\cite{sun2023samwsss, jiang2023sam-pseudo-labeler, chen2023sepl} utilize SAM to deal with the weakly-supervised semantic segmentation problem. However, SAM is an interactive segmentation method, which heavily relies on human prompts.

In our approach, we innovatively propose to automatically prompt SAM using classification clues to extract region proposals. This method results in high-recall proposals that surpass traditional methods like Selective Search in terms of both efficiency and effectiveness. This advancement represents a significant improvement in the domain of proposal generation within the WSOD framework.

\vspace{-0.1 in}

\begin{figure*}[t]
    \centering
    \includegraphics[width=1.0\linewidth]{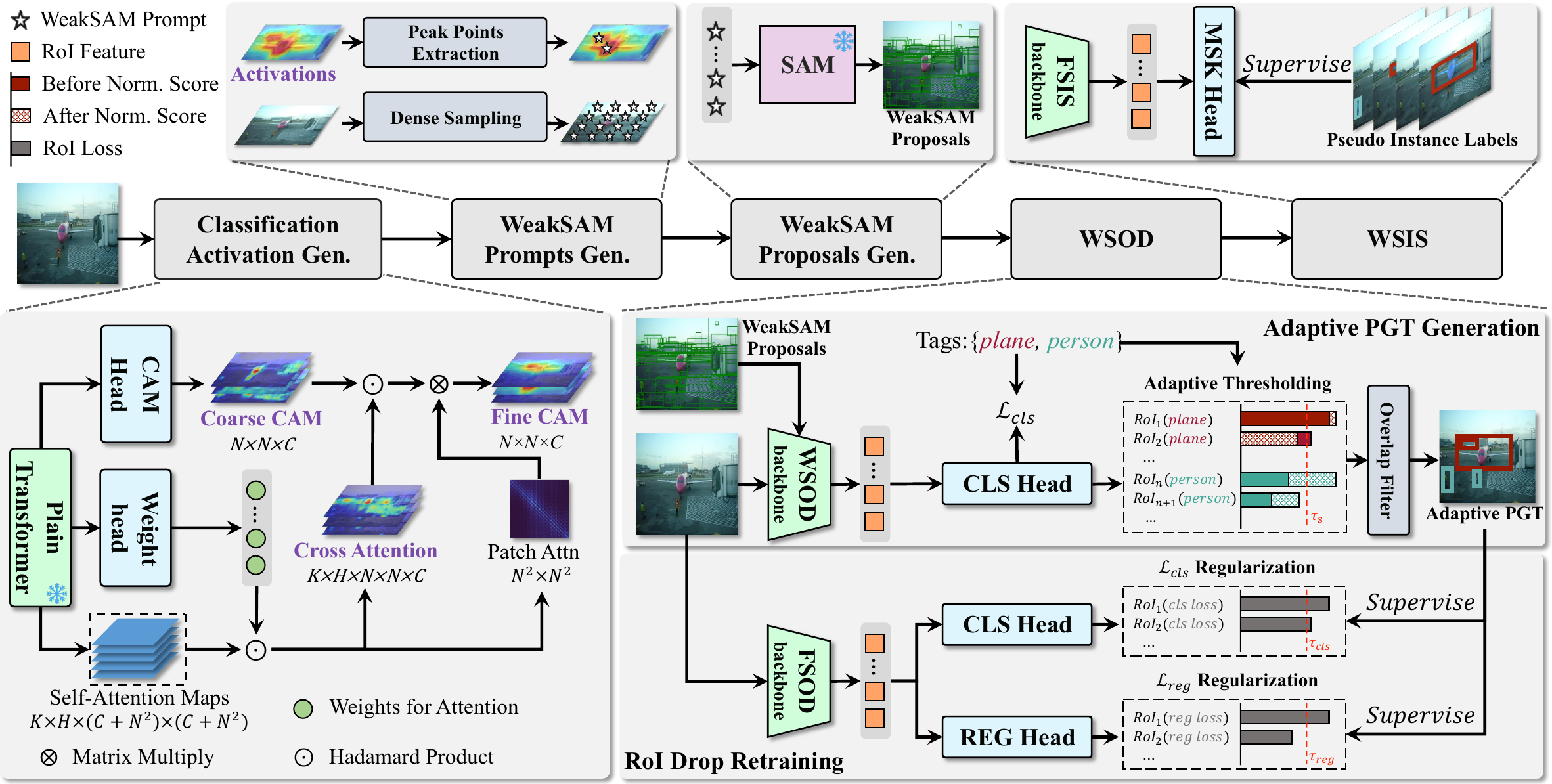}
    \vspace{-0.2 in}
    \caption{An overview of the proposed \name{} framework. 
    We first generate activation maps from a classification ViT~\cite{zhu2023weaktr}. 
    Subsequently, we introduce classification clues and spatial points as automatic \name{}~prompts, which address the problem of SAM requiring interactive prompts.
    Next, we use the \name{} proposals in the WSOD pipeline, in which the weakly-supervised detector performs class-aware perception to annotate pseudo ground truth (PGT).
    Then, we analyze the incompleteness and the noise problem existing in PGT and propose adaptive PGT generation, RoI drop regularization to address them, respectively.
    Finally, we launch WSIS training supervised by pseudo instance labels, which requires adaptive PGT as SAM prompts.
    The snowflake mark means the model is frozen. %
    }
    \label{fig:pipeline}
\vspace{-0.15 in}
\end{figure*}

\subsection{Weakly-supervised Object Detection}
Weakly-supervised object detection (WSOD) with image-level labels~\cite{laptevcontextlocnet, diba2017weakly, tang2018weakly, gao2018c, wan2018min, zhang2018zigzag, liu2019utilizing, Li_2019_ICCV, Arun_2019_CVPR, sun2020mining, arun2020weakly, jia2021gradingnet, wan2019c} is important for reducing the human annotation burden. The previous works, i.e., WSDDN~\cite{bilen2016wsddn} and OICR~\cite{tang2017multiple}, proposed the Multiple Instance Learning and online refinement paradigms. The later works aimed to improve the WSOD performance from different perspectives. Such as WSOD$^2$~\cite{zeng2019wsod2} introduced bottom-up object evidence, PCL~\cite{tang2018pcl} proposed to cluster proposals, MIST~\cite{ren2020instance} utilized a self-training algorithm, etc. 
\color{black}
Besides, some methods~\cite{tang2018pcl, jie2017self-taught, li2016pda, sui2022sos-wsods, zhang2018w2f, huang2022w2nswitching} also retrained a fully-supervised object detection network with generated pseudo ground truth (PGT). However, most of them used the proposals generated from low-level methods, i.e., Selective Search~\cite{Uijlings2013SelectiveSF}, EdgeBox~\cite{zitnick2014eb}, and MCG~\cite{pont2016mcg}, which contain a great number of redundant proposals and bring an optimization challenge.

Different from previous methods, our \name{} proposals have fewer numbers and higher recall, which reduces the difficulty of finding the correct proposals for WSOD methods. For the key problem of PGT incompleteness and noisy PGT instances, we propose adaptive PGT generation and Region of Interest (RoI) drop regularization to address them, respectively. 

\vspace{-0.1 in}

\subsection{Weakly-supervised Instance Segmentation}
Weakly-supervised instance segmentation (WSIS) aims to achieve instance segmentation through weak supervision, such as box-level supervision~\cite{tian2021boxinst, wang2021boxcaseg, cheng2023boxteacher, hsu2019bbtp, liao2019weakly, Lee_2021_CVPR, Khoreva_2017_CVPR, zhang2023weakly, zhu2023encoder, li2022box}, and image-level supervision~\cite{ge2019label-penet, ou2021wsrcnn, Zhu2019iam,liu2020liid, Hwang_2021_WACV, zhang2021weakly, hu2020weakly, Hsieh_2023_ICCV,laradji2019masks}. The WSIS with image-level supervision is challenging because it lacks accurate instance locations. Some image-level WSIS methods use class activation map (CAM)~\cite{zhou2015learning} to extract coarse object locations, such as PRM~\cite{zhou2018prm}, IAM~\cite{Zhu2019iam}, IRNet~\cite{Ahn2019irn}, BESTIE~\cite{Kim2022bestie}, etc. Some other image-level WSIS methods try to incorporate instance clues from extra priors, such as Fan et al.~\cite{fan2018wsss}, LIID~\cite{liu2020liid}, CIM~\cite{zecheng2023cim}, etc. However, they always need complicated networks and lack high-quality instance segments.

Different from previous WSIS methods, the proposed WSIS extension using \name{} PGT and SAM's prediction is concise and effective. The generated pseudo instance labels can further be applied to any fully-supervised instance segmentation method.

\section{Methods}
\label{sec:methods}
We present the \name{} framework as shown in Fig.~\ref{fig:pipeline}.
At first, \name{} collects classification activations from a classification ViT.
Subsequently, \name{} automatically generates prompts from classification activations and spatial samples. Next, \name{} sends the prompts to SAM for \name{} proposals. 
Then, we launch the weakly-supervised object detection (WSOD) pipeline, which is enhanced by \name{} proposals, adaptive pseudo ground truth (PGT) generation, and RoI drop regularization. 
Last, we use the SAM-enhanced pseudo instance labels to launch the weakly-supervised instance segmentation extension.

\subsection{Classification Clues as Automatic Prompts}

Previous WSOD methods face an optimization problem caused by the redundant proposals, e.g., Selective Search~\cite{Uijlings2013SelectiveSF} and EdgeBox~\cite{zitnick2014eb}, because these proposals are only based on low-level features.
To address this problem, we propose to transfer knowledge in the foundation model, i.e., SAM, for proposal generation. We use classification clues to prompt SAM automatically, which also solves the shortcoming of SAM requiring interactive prompts.

\vspace{-.08in}

\paragraph{Classification Activation Generation}

As shown in Fig.~\ref{fig:pipeline}, we extract classification clues from a classification ViT. 
Specifically, we choose the pre-trained weakly-supervised semantic segmentation network, WeakTr~\cite{zhu2023weaktr}, to provide classification clues because of its superior localization ability. 
At first, we extract cross-attention maps $ \mathbf{CA} \in \mathbb{R} ^ {K \times H \times N \times N \times C} $ from the self-attention maps, where $K$ is the number of transformer encoding layers, $H$ is the number of attention heads in each layer, $N \times N$ is the spatial size of the visual tokens, and $C$ represents the total number of classification categories. Then, we obtain coarse class activation map (CAM)~\cite{zhou2015learning}, $\mathbf{CAM}_{coarse} \in \mathbb{R} ^ {N \times N \times C}$, from the convolutional CAM head, which takes visual tokens at the final transformer layer as input and produces coarse CAM. 
Last, we use coarse CAM and weighted self-attention maps to produce fine CAM, $\mathbf{CAM}_{fine} \in \mathbb{R} ^ {N \times N \times C}$. 

\vspace{-.08 in}

\paragraph{\name{}~Prompts Generation}

As shown in Fig.~\ref{fig:pipeline}, we extract prompts from dense sampling points and activations, which include cross-attention maps, coarse CAM, and fine CAM. 
At first, the dense sampling requires splitting the image into $ S \times S $ patches and taking the center points as prompts. Notably, the dense sampling points provide spatial-aware prompts but lack explicit reference to objects and semantics.
Then, we get peak points from the cross-attention maps as prompts. We observe that these maps do not solely concentrate on objects from their corresponding categories but also give attention to objects from different categories. So, we mark these prompts as instance-aware ones. 
Last, we extract peak points from coarse CAM and fine CAM as semantic-aware prompts, which are more precise and focus on areas of foreground objects.

Specifically, we extract peak points from cross-attention maps and CAMs, as shown in Algorithm~\ref{alg:peak_points_extraction}. 
Given cross-attention maps or CAMs as input, we first initialize the peak points list $P$, peak values list $V$, deleted lists $P_{\mathrm{delete}}$, $V_{\mathrm{delete}}$, and max pooling operation. 
Next, we reshape the input maps and ensure the last two dimensions correspond to the original image size and the others as the first dimension. 
Then, we apply max pooling on the input maps $M$, and sort $V$ and $P$ in descending order based on $V$. Last, we remove points with low activation values or close to high-score points.

\vspace{-.08 in}

\paragraph{\name{} Proposals Generation}

At the \name{} proposal generation stage, we use the three kinds of prompts to prompt SAM automatically. 
We directly add semantic-aware prompts and spatial-ware prompts to the prompt list, because they usually have clear localization to foreground objects and spatial positions, respectively. 
For the instance-aware prompts that have some redundancy, we cluster them to filter the duplicated ones and then add them to the prompt list. 
Finally, the prompt list is used to prompt SAM for \name{} proposals.

\vspace{-.1 in}

{
\begin{algorithm}[htb]
\caption{Peak Points Extraction}\label{alg:peak_points_extraction}
\small
\begin{algorithmic}[1]
\REQUIRE
{maps $\mathbf{M}$ ($\mathbf{CA}$ or $\mathbf{CAM}$), kernel size $k$, activation threshold $\tau$}
\ENSURE
{ peak points coordinates list $P = [ p_0, p_1, \dots, p_{n-1}]$, corresponding peak values list $V = [ v_0, v_1, \dots, v_{n-1}]$}
\STATE $\mathbf{M}$ = $\mathbf{M}$.$\mathrm{view}$(-1, $N$, $N$) \textcolor{gray}{\text{// reshape }}
\STATE Initialize $P$, $V$ as empty list
\STATE Initialize $\mathrm{Maxpool()}$ operation with kernel size $k$
\STATE $P$, $V$ =  $\mathrm{Maxpool(\mathbf{M})}$ \textcolor{gray}{\text{// get coordinates and values}}
\STATE Sort $V$ in descending order of numerical value, and rearrange $P$ accordingly
\STATE Initialize list $P_{\text{delete}}$, $V_{\text{delete}}$ to mark points for deletion
\FOR{each index $i$ from $0$ to $\mathrm{length}(P)$}
\STATE \textcolor{gray}{\text{// skip further checks for points marked for deletion}}
\IF{$p_i$ in $P_{\mathrm{delete}}$}
\STATE Continue
\ENDIF
\STATE \textcolor{gray}{\text{// mark activation points with low score}}
\IF{$v_i  <  \tau $}
\STATE Append $p_i$, $v_i$ to $P_{\text{delete}}$, $V_{\text{delete}}$
\STATE Continue
\ENDIF
\STATE \textcolor{gray}{\text{// mark lower-score points near the current point}}
\FOR{each index $j = i+1$ to $\mathrm{length}(P)$}
\IF{$\lvert \lvert p_j - p_i\rvert \rvert \leq k / 2$}
\STATE Append $p_j$, $v_j$ to $P_{\text{delete}}$, $V_{\text{delete}}$
\ENDIF
\ENDFOR
\ENDFOR
\STATE Remove all points in $P_{\text{delete}}$ and $V_{\text{delete}}$ from $P$ and $V$
\STATE \textbf{return $P$, $V$}
\end{algorithmic}
\end{algorithm}
}

\vspace{-.1in}

\subsection{\name{} WSOD Pipeline}

To better describe the proposed weakly-supervised object detection (WSOD) pipeline, we first present the weakly-supervised detector training with \name{} proposals. Then, we identify the PGT incompleteness problem and introduce the proposed adaptive PGT generation to address it. Last, we analyze the noise problem existing in the retraining phase, and propose Region of Interest (RoI) drop regularization to alleviate the effect of noise.

\vspace{-0.08 in}

\paragraph{Weakly-supervised Detector Training}

A primary challenge in traditional WSOD methods is the low training efficiency, largely attributed to the redundancy of proposals. Traditional approaches often involve the Region of Interest pooling layer processing thousands of proposals per image, which impairs both effectiveness and efficiency. To address this issue, our \name{} proposals adopt transferred knowledge from SAM and classification clues. The proposed method focuses on generating a smaller quantity of proposals while maintaining high recall, thereby enhancing the overall efficiency and efficacy of the detection process in a WSOD context.
We apply the proposed \name{} on previous WSOD methods, including OICR~\cite{tang2017multiple} and MIST~\cite{ren2020instance}, which receive significant improvements. 

\vspace{-0.08 in}

\paragraph{Adaptive PGT Generation}

Generating high-quality pseudo ground truth (PGT) is the key to the WSOD paradigm. 
Traditional WSOD methods often encounter the issue of PGT incompleteness. This occurs because these methods typically select top-scoring proposals as PGT or apply a uniform threshold to filter proposals across all categories. Such approaches can lead to the omission of objects or entire categories, especially when proposals in certain categories score low. 
To address these problems, we propose an adaptive PGT generation method to normalize the score distribution of proposals, ensuring they fall within a similar range, as shown in Algorithm.~\ref{alg:adaptive_pgt_generation}.

For box list $B \in \mathbb{R}^{N \times 5}$ and corresponding score list $S \in \mathbb{R}^{N \times 1}$, we first select them with a specific classification label and then normalize the scores. The $N$ is the number of predicted boxes, and the second dimension of $B$ is the combination of a category label and four coordinate values.
Next, we keep boxes with scores higher than the threshold $\tau_s$. Please note that the normalization enables the threshold to work for all categories adaptively, so we would not lose a ground truth category even if all boxes in this category have low scores. Then, we select the boxes whose main parts are not contained in some bigger boxes. Because the boxes that have more $overlap$ are often local components of some objects. Last, we return the box list $B'$ as the final PGT.

\vspace{-.1in}

{
\begin{algorithm}[ht]
\caption{Adaptive Pseudo Ground Truth Generation}\label{alg:adaptive_pgt_generation}
\small

\begin{algorithmic}[1]
\REQUIRE
{boxes list $B$ of an image, corresponding scores list $S$, corresponding classification labels $Y$, score threshold $\tau_s$, overlap threshold $\tau_o$}
\ENSURE
{ pseudo ground truth boxes $B'$}
\STATE initialize $B'$ as empty list
\FOR{each $y_i$ in $Y$}
\STATE \textcolor{gray}{\text{// get boxes' indices with label $y_i$}}
\STATE $idx_i$ = $\mathrm{where}$ ($B[:, 0]$ == $y_i$ ) 
\STATE $S_i$ = $S[idx_i, :]$  
\STATE $B_i$ = $B[idx_i, :]$  
\STATE $S_i^{norm}$ =  $\frac{S_i - \mathrm{min}(S_i)}{\mathrm{max}(S_i) - \mathrm{min}(S_i)}$ \text{// normalize scores}
\STATE \textcolor{gray}{\text{// keep boxes with high score}}
\STATE $idx_{keep}$ = $\{ j \ | \ s_j \in S_i^{norm}, \ s_j > \tau_s \}$
\STATE $B_i$ = $B_i[idx_{keep}, :]$  
\STATE $S_i^{norm}$ = $S_i^{norm}[idx_{keep}, :]$  
\STATE \textcolor{gray}{\text{// select boxes with less overlap}}
\FOR{each box $b_j$ in $B_i$}
\STATE $overlaps$ = $\{\frac{|b_j \cap b_k|}{|b_j|} \ | \ b_k \in B_i, \ k \neq j\}$
\IF{all $overlap < \tau_o$ in $overlaps$}
\STATE Append $b_j$ to $B'$
\ENDIF
\ENDFOR
\ENDFOR
\STATE \textbf{return $B'$}

\end{algorithmic}
\end{algorithm}
}

\vspace{-.2in}

\paragraph{RoI Drop Regularization}

\begin{figure}[htp]
  \vspace{-0.2 in}
  \centering
  \setlength{\tabcolsep}{1.6 pt}
  \includegraphics[width=1.\linewidth]{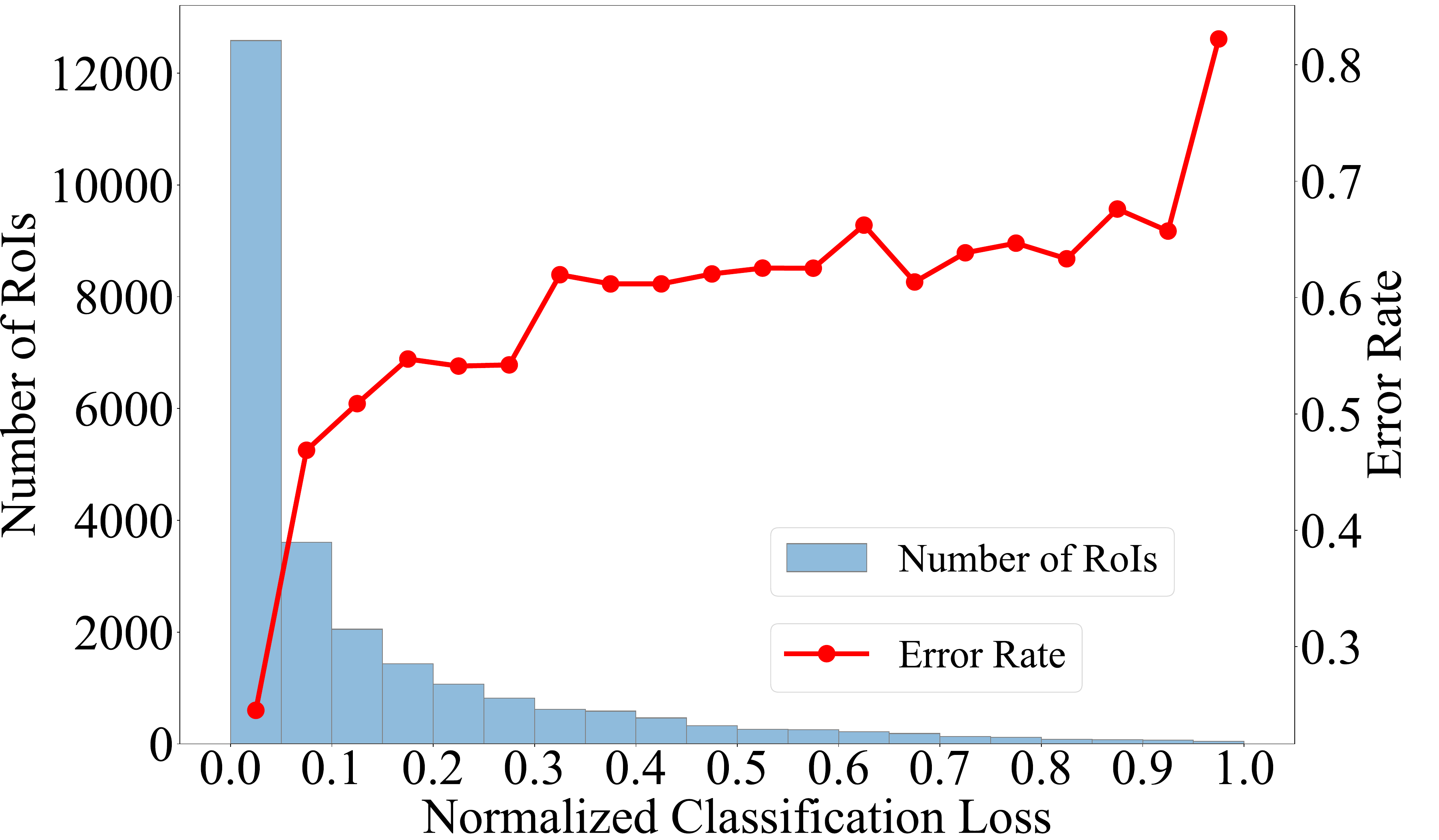}
  \vspace{-0.3in}
  \caption{
  The relationship between the normalized classification loss, corresponding number of RoIs and error rate. The results are obtained from training the Faster-RCNN using PGT in the preliminary training stage.}
   \label{fig:loss_drop}
   \vspace{-0.2in}
\end{figure}

A recognized issue in the retraining phase of WSOD is noisy PGT instances. 
These noisy instances result in PGT acting as the inaccurate supervision.
Alleviating this problem is critical for enhancing the performance of WSOD retraining.
To analyze this problem in depth, we first divide the RoIs into different loss intervals. Then, we mark the RoIs whose corresponding PGTs do not have at least 70\% IoU with the ground truth boxes as error ones. Last, we present the statistics as shown in Fig.~\ref{fig:loss_drop}, which demonstrates that the RoIs with larger losses are in a small amount and have a high error rate.

Intuitively, we propose a method, named RoI drop regularization, to adaptively drop the RoIs with larger losses. Notably, the proposed method is easy to implement and can further help the query-based detectors to alleviate the noisy PGT problem by its variant, query drop regularization. For anchor-based FSOD methods, e.g., Faster-RCNN~\cite{ren2015faster}, we first determine the thresholds $\tau_{cls}$ and $\tau_{reg}$ for classification loss and regression loss, respectively. Then, we compute the drop signal $d_i$ for $i\mathrm{-th}$ RoI.

\vspace{-.1 in}

\begin{align}
{d}_i &=
  \left
  \{\begin{array}{ll}
  1, &  l_{i}^{cls} \le \tau_{cls},~\mathrm{and}~l_{i}^{reg} \le \tau_{reg} \\
  0, &  \mathrm{others}
  \end{array} 
  \right. ,
\vspace{-.4 in}
\end{align}
where the $l_{i}^{cls}$ and $l_{i}^{reg}$ represent the classification loss and regression loss for each RoI, respectively. When the two losses of a RoI are all below their thresholds, we set its drop signal $d_i$ as $1$. Finally, we integrate the $d_i$ into the computation of final loss $\mathcal{L}$.

\vspace{-0.1 in}

\begin{align}
    \mathcal{L} = \sum_{i} d_i l^{cls}_i  
    + \lambda \sum_{i} p^{*}_i d_i l^{reg}_i ,
\end{align}
where $p^*_i$ is 1 if the box is positive, and 0 if the box is negative. The $\lambda$ is a balancing weight.

For query-based FSOD methods, e.g., DINO~\cite{zhang2022dino}, since queries can be regarded as dynamic RoIs, we apply query drop regularization on them. Because only a few matched queries need to calculate box loss $l^{box}$ and IoU loss $l^{iou}$, we only set a percentile threshold based on classification loss $l^{cls}$.
Only when the $i\mathrm{-th}$ query's loss $l^{cls}_i$ is less than the loss at $\tau$\% percentile, i.e., $l_{\tau}^{cls}$, will its corresponding $d_i$ be set to $1$.

\vspace{-0.2 in}

\begin{align}
{d}_i &=
  \left
  \{\begin{array}{ll}
  1, &  l_{i}^{cls} \le l_{\tau}^{cls} \\
  0, &  \mathrm{others}
  \end{array} 
  \right. .
\end{align}

\vspace{-0.2 in}

\begin{align}
    \mathcal{L}_\mathrm{Hungarian} = \sum_{i} d_i[l^{cls}_i + p^*_il^{box}_i + p^*_il^{iou}_i] .
\end{align}

\begin{table*}[ht]
\centering
\small
\renewcommand\arraystretch{.85}
\caption{
Comparisons of the WSOD performance in terms of $\text{AP}$ metrics on three benchmarks: PASCAL VOC 2007, PASCAL VOC 2012, and COCO 2014. The $Sup.$ column denotes the type of supervision used for training including full supervision ($\mathcal{F}$), point-level labels ($\mathcal{P}$), image-level labels ($\mathcal{I}$). ``*'' means the results rely on MCG~\cite{pont2016mcg} proposals. ``\ddag'' means this method use the a heavy RN50-WS-MRRP~\cite{shen2020enabling} backbone (1.76 $\times$ parameters than VGG16 and 10.10 $\times$ parameters than RN50). We mark the best WSOD results in bold.
}   %
\vspace{-0.08 in}
\begin{tabular}{lccccccccccccccc}
\toprule
                \multicolumn{2}{l}{\multirow{2}{*}{\textbf{Methods}}}                       & \multicolumn{2}{c}{\multirow{2}{*}{\textbf{Proposal}}} & \multirow{2}{*}{$Sup.$} & \multirow{2}{*}{\textbf{Retrain}}
& \multicolumn{2}{c}{\textbf{VOC 07}} & \multicolumn{2}{c}{\textbf{VOC 12}} &            \multicolumn{6}{c}{\textbf{COCO 14}}\\
 \multicolumn{2}{c}{}& \multicolumn{2}{c}{}& & & \multicolumn{2}{c}{$\text{AP}_{\scriptsize50}$}& \multicolumn{2}{c}{$\text{AP}_{\scriptsize50}$} & \multicolumn{2}{c}{$\text{AP}_{\scriptsize{50:95}}$}& \multicolumn{2}{c}{$\text{AP}_{\scriptsize50}$}& \multicolumn{2}{c}{$\text{AP}_{\scriptsize75}$}\\
\midrule
\multicolumn{8}{l}{\textbf{\textit{Fully-supervised object detection methods.}}} \\
 \multicolumn{2}{l}{Faster R-CNN~\cite{ren2015faster}}       & \multicolumn{2}{c}{\multirow{1}{*}{RPN}} &            \multirow{1}{*}{$\mathcal{F}$}     & --
& \multicolumn{2}{c}{69.9}      & \multicolumn{2}{c}{--} &       \multicolumn{2}{c}{21.2}& \multicolumn{2}{c}{41.5}& \multicolumn{2}{c}{--}\\
\midrule
\multicolumn{8}{l}{\textbf{\textit{WSOD methods with point supervision.}}} \\
 \multicolumn{2}{l}{P2BNet~\cite{chen2022point}}                       & \multicolumn{2}{c}{RPN} &       $\mathcal{P}$     & --
& \multicolumn{2}{c}{60.2}    & \multicolumn{2}{c}{--} &         \multicolumn{2}{c}{19.4}& \multicolumn{2}{c}{43.5}& \multicolumn{2}{c}{--}\\
\midrule
\multicolumn{8}{l}{\textbf{\textit{WSOD methods with image-level supervision.}}} \\
                \multicolumn{2}{l}{C-MIDN                        ~\cite{gao2019c}}              & \multicolumn{2}{c}{SS, MCG} &  \multirow{9}{*}{$\mathcal{I}$}     & --  
& \multicolumn{2}{c}{52.6}      & \multicolumn{2}{c}{50.2} &       \multicolumn{2}{c}{9.6$^*$}& \multicolumn{2}{c}{21.4$^*$
}& \multicolumn{2}{c}{--
}\\

\multicolumn{2}{l}{WSOD$^2$                        ~\cite{zeng2019wsod2}}              & \multicolumn{2}{c}{SS} &       & --
& \multicolumn{2}{c}{53.6}      & \multicolumn{2}{c}{47.2} &       \multicolumn{2}{c}{10.8}& \multicolumn{2}{c}{22.7
}& \multicolumn{2}{c}{--
}\\
\multicolumn{2}{l}{SLV                       ~\cite{Chen_2020_CVPR}}              & \multicolumn{2}{c}{SS} &       & --
& \multicolumn{2}{c}{53.5}      & \multicolumn{2}{c}{49.2} &       \multicolumn{2}{c}{--}& \multicolumn{2}{c}{--
}& \multicolumn{2}{c}{--
}\\

                \multicolumn{2}{l}{CASD                          ~\cite{huang2020comprehensive}} & \multicolumn{2}{c}{SS} &       & --
& \multicolumn{2}{c}{56.8}     & \multicolumn{2}{c}{53.6} &        \multicolumn{2}{c}{12.8}& \multicolumn{2}{c}{26.4}& \multicolumn{2}{c}{--
}\\

                \multicolumn{2}{l}{IM-CFB                          ~\cite{yin2021instance}} & \multicolumn{2}{c}{SS} &       & --
& \multicolumn{2}{c}{54.3}     & \multicolumn{2}{c}{49.4} &        \multicolumn{2}{c}{--}& \multicolumn{2}{c}{--}& \multicolumn{2}{c}{--
}\\

                \multicolumn{2}{l}{OD-WSCL                       ~\cite{seo2022object}}         & \multicolumn{2}{c}{SS, MCG} &       & --
& \multicolumn{2}{c}{56.4}     & \multicolumn{2}{c}{54.6} &        \multicolumn{2}{c}{13.7$^*$}& \multicolumn{2}{c}{27.7$^*$}& \multicolumn{2}{c}{11.9$^*$
}\\
                \multicolumn{2}{l}{WSOD-CBL                      ~\cite{yin2023cyclic}}        & \multicolumn{2}{c}{SS} &       & --
& \multicolumn{2}{c}{57.4}     & \multicolumn{2}{c}{53.5} &        \multicolumn{2}{c}{13.6}& \multicolumn{2}{c}{27.6}& \multicolumn{2}{c}{--}\\
\multicolumn{2}{l}{WSOVOD                      ~\cite{lin2023weakly}} & \multicolumn{2}{c}{LO-WSRPN + SAM} & & -- & \multicolumn{2}{c}{59.1}     & \multicolumn{2}{c}{59.8} &        \multicolumn{2}{c}{18.8}& \multicolumn{2}{c}{27.1}& \multicolumn{2}{c}{19.7}\\
\multicolumn{2}{l}{WSOVOD\ddag                } & \multicolumn{2}{c}{LO-WSRPN + SAM} & & -- & \multicolumn{2}{c}{63.4}     & \multicolumn{2}{c}{62.1} &        \multicolumn{2}{c}{20.5}& \multicolumn{2}{c}{29.1}& \multicolumn{2}{c}{21.4}\\

\midrule
\multicolumn{8}{l}{\textbf{\textit{Baseline and ours.}}} \\
\multicolumn{2}{l}{OICR~\cite{tang2017multiple}}      & \multicolumn{2}{c}{SS, MCG} &   \multirow{2}{*}{$\mathcal{I}$}  &  --  & \multicolumn{2}{c}{41.2}     & \multicolumn{2}{c}{37.9} &  \multicolumn{2}{c}{8.0$^*$}& \multicolumn{2}{c}{18.9$^*$}& \multicolumn{2}{c}{7.0$^*$
}\\

                \multicolumn{2}{l}{\name{} (OICR)}                             & \multicolumn{2}{c}{\name{}} &      & -- & \multicolumn{2}{c}{58.9\scriptsize\color{red}{+17.7}}      & \multicolumn{2}{c}{58.4\scriptsize\color{red}{+20.5}} &  \multicolumn{2}{c}{19.9\scriptsize\color{red}+11.9}& \multicolumn{2}{c}{32.1\scriptsize\color{red}+13.2}& \multicolumn{2}{c}{20.6\scriptsize\color{red}+13.6}\\
\midrule
\multicolumn{8}{l}{\textbf{\textit{Baseline and ours.}}} \\
 \multicolumn{2}{l}{MIST         ~\cite{ren2020instance}}       & \multicolumn{2}{c}{SS, MCG} &                       \multirow{2}{*}{$\mathcal{I}$} & --
& \multicolumn{2}{c}{54.9}  & \multicolumn{2}{c}{52.1} &  \multicolumn{2}{c}{11.4$^*$}& \multicolumn{2}{c}{24.3$^*$}& \multicolumn{2}{c}{9.4$^*$
}\\
                \multicolumn{2}{l}{\name{} (MIST)}                             & \multicolumn{2}{c}{\name{}} &   & --
& \multicolumn{2}{c}{67.4\scriptsize\color{red}{+12.5}}   & \multicolumn{2}{c}{66.9\scriptsize\color{red}{+14.8}} &    \multicolumn{2}{c}{22.9\scriptsize\color{red}+11.5}& \multicolumn{2}{c}{35.2\scriptsize\color{red}+10.9}& \multicolumn{2}{c}{24.6\scriptsize\color{red}+15.2}\\
\midrule
\multicolumn{8}{l}{\textbf{\textit{WSOD methods with image-level supervision. + Retrain}}} \\
 \multicolumn{2}{l}{W2F~\cite{zhang2018w2f}}& \multicolumn{2}{c}{RPN}& \multirow{3}{*}{$\mathcal{I}$}& Faster R-CNN &\multicolumn{2}{c}{52.4}& \multicolumn{2}{c}{47.8}& \multicolumn{2}{c}{--
}& \multicolumn{2}{c}{--}& \multicolumn{2}{c}{--}\\
 \multicolumn{2}{l}{SoS-WSOD~\cite{sui2022sos-wsods}}& \multicolumn{2}{c}{RPN}& & Faster R-CNN & \multicolumn{2}{c}{64.4}& \multicolumn{2}{c}{61.9}& \multicolumn{2}{c}{16.6}& \multicolumn{2}{c}{32.8}& \multicolumn{2}{c}{15.2}\\
 \multicolumn{2}{l}{W2N~\cite{huang2022w2nswitching} }& \multicolumn{2}{c}{RPN}& & Faster R-CNN & \multicolumn{2}{c}{65.4}& \multicolumn{2}{c}{60.8}& \multicolumn{2}{c}{15.9}& \multicolumn{2}{c}{33.3}& \multicolumn{2}{c}{13.4}\\
 \midrule
 \multicolumn{8}{l}{\textbf{\textit{Ours. + Retrain}}} \\
  \multicolumn{2}{l}{\name{} (OICR)}& \multicolumn{2}{c}{RPN}& \multirow{4}{*}{$\mathcal{I}$} & Faster R-CNN & \multicolumn{2}{c}{65.7}& \multicolumn{2}{c}{62.9}& \multicolumn{2}{c}{22.3}& \multicolumn{2}{c}{36.5}& \multicolumn{2}{c}{23.0}\\
  \multicolumn{2}{l}{\name{} (MIST)}& \multicolumn{2}{c}{RPN}&   & Faster R-CNN & \multicolumn{2}{c}{71.8}& \multicolumn{2}{c}{69.2}& \multicolumn{2}{c}{23.8}& \multicolumn{2}{c}{38.5}& \multicolumn{2}{c}{25.1}\\
  \multicolumn{2}{l}{\name{} (OICR)}& \multicolumn{2}{c}{--}&   & DINO & \multicolumn{2}{c}{66.1}& \multicolumn{2}{c}{63.7}& \multicolumn{2}{c}{24.9}& \multicolumn{2}{c}{36.9}& \multicolumn{2}{c}{26.8}\\ %
 \multicolumn{2}{l}{\name{} (MIST)}& \multicolumn{2}{c}{--}&   & DINO & \multicolumn{2}{c}{\textbf{73.4}}& \multicolumn{2}{c}{\textbf{70.2}}& \multicolumn{2}{c}{\textbf{26.6}}& \multicolumn{2}{c}{\textbf{39.3}}& \multicolumn{2}{c}{\textbf{29.0}}\\

\toprule
\end{tabular}

\label{tab:perf_Wsod}
\vspace{-0.05 in}
\end{table*}

\vspace{-.05in}

\subsection{\name{} for WSIS}

\begin{table*}[ht]
\centering
\renewcommand\arraystretch{.85}
\caption{
Comparisons of the WSIS performance in terms of $\text{AP}$ metrics on PASCAL VOC 2012. The $Sup.$ column denotes the type of supervision used for training including mask supervision ($\mathcal{M}$), saliency maps ($\mathcal{S}$), image-level labels ($\mathcal{I}$),  and SAM models ($\mathcal{A}$). We mark the best WSIS results in bold.
}
\vspace{-0.08in}
\begin{tabular}{lccccccccc}
\toprule
                \multicolumn{2}{l}{\multirow{2}{*}{\textbf{Methods}}}                       & \multicolumn{2}{c}{\multirow{2}{*}{\textbf{Backbone}}} & \multirow{2}{*}{$Sup.$} & \multirow{2}{*}{\textbf{Retrain}}
& \multicolumn{4}{c}{\textbf{VOC 12}}\\
 \multicolumn{2}{c}{}& \multicolumn{2}{c}{}& & & $\text{AP}_{\scriptsize25}$&$\text{AP}_{\scriptsize50}$& $\text{AP}_{\scriptsize70}$& $\text{AP}_{\scriptsize75}$\\
\midrule
\multicolumn{8}{l}{\textbf{\textit{Fully-supervised instance segmentation methods.}}} \\
 \multicolumn{2}{l}{Mask R-CNN~\cite{he2017mask}}       & \multicolumn{2}{c}{ResNet-101} &            $\mathcal{M}$& -- & 76.7&      67.9& 52.5& 44.9\\ %
\midrule
\multicolumn{8}{l}{\textbf{\textit{WSIS methods with image-level supervision. + Retrain}}} \\
                \multicolumn{2}{l}{WISE~\cite{laradji2019masks}} & \multicolumn{2}{c}{ResNet-50} &        
$\mathcal{I}$& Mask R-CNN & 49.2&     41.7& --& 23.7\\
                \multicolumn{2}{l}{IRNet~\cite{Ahn2019irn}}         & \multicolumn{2}{c}{ResNet-50} &        
$\mathcal{I}$& Mask R-CNN & --&     46.7& 23.5& --\\
                \multicolumn{2}{l}{LIID~\cite{liu2020liid}}        & \multicolumn{2}{c}{ResNet-50} &        
$\mathcal{I+S}$& Mask R-CNN & --&     48.4& --& 24.9\\
 \multicolumn{2}{l}{Arun et al.\cite{arun2020weakly}}& \multicolumn{2}{c}{ResNet-50}& $\mathcal{I}$& Mask R-CNN & 59.7&50.9& 30.2& 28.5\\
 \multicolumn{2}{l}{WS-RCNN~\cite{ou2021wsrcnn}}& \multicolumn{2}{c}{VGG-16}& $\mathcal{I}$& Mask R-CNN & 62.2&47.3& --& 19.8\\
\multicolumn{2}{l}{BESTIE~\cite{Kim2022bestie}}      & \multicolumn{2}{c}{HRNet-W48} &   $\mathcal{I}$      & Mask R-CNN & 61.2&     51.0& 31.9& 26.6\\

 \multicolumn{2}{l}{CIM~\cite{zecheng2023cim}}& \multicolumn{2}{c}{ResNet-50}& $\mathcal{I}$  & Mask R-CNN    & 68.7& 55.9& 37.1& 30.9\\
 \midrule
 \multicolumn{8}{l}{\textbf{\textit{Ours.}}} \\
  \multicolumn{2}{l}{\name{}}& \multicolumn{2}{c}{ResNet-50}& $\mathcal{I + A}$      & Mask R-CNN & 70.3 & 59.6 & 43.1 & 36.2 \\
  \multicolumn{2}{l}{\name{}}& \multicolumn{2}{c}{ResNet-50}& $\mathcal{I + A}$      & Mask2Former & \textbf{73.4} & \textbf{64.4} & \textbf{49.7} & \textbf{45.3}\\
 \bottomrule
\end{tabular}

\label{tab:perf_wsis_voc}
\vspace{-0.05 in}
\end{table*}

Thanks to the high-quality \name{} PGT, we can directly use them to prompt SAM for precise segments as pseudo instance labels. Following the practices in the \name{} WSOD pipeline, we evaluate the quality of \name{} PGT using R-CNN-based and query-based instance segmentation methods, respectively. Notably, we do not introduce more techniques in the \name{} WSIS, because the \name{} pseudo instance labels are accurate enough.

\section{Experiment}
\label{sec:exps}

\subsection{Experimental Setup}
\paragraph{Datasets and Metrics}
We evaluate the proposed \name{} on both weakly-supervised object detection (WSOD) and weakly-supervised instance segmentation (WSIS) benchmarks. Notably, the same datasets for different tasks may have different settings. For \textbf{WSOD}, we use three datasets, i.e., PASCAL VOC 2007~\cite{everingham2015voc}, PASCAL VOC 2012~\cite{everingham2015voc}, and COCO 2014~\cite{lin2014microsoft}. PASCAL VOC 2007 has 2501 images for training, 2510 images for evaluation, and 4592 images for testing. PASCAL VOC 2012 contains 5717 training images, 5823 validation images, and 10991 test images. COCO 2014 includes around 80,000 images for training and 40,000 images for validation. Following previous WSOD methods, we train \name{} on $train$ and $val$ sets and evaluate \name{} on the $test$ set for PASCAL VOC 2007 and 2012. For COCO 2014, we use the $train$ set for training and the $val$ set for evaluating. 
PASCAL VOC 2007 and 2012 datasets comprise 20 object categories and COCO 2014 comprises 80 ones. We report the average precision \text{AP} metrics for these benchmarks.  
For \textbf{WSIS}, we use two datasets, i.e., PASCAL VOC 2012, and COCO 2017. The PASCAL VOC 2012 dataset includes 10582 images for training, and 1449 images for evaluation, comprising 20 object categories. The COCO 2017 dataset includes 115K training images, 5K validation images, and 20K testing images, comprising 80 object categories. Following previous methods, we report the average precision \text{AP} metrics with different Intersection-over-Union (IoU) thresholds. %

\begin{table*}[htp]
\small
\centering
\renewcommand\arraystretch{1}
\caption{
Comparisons of the WSIS performance in terms of $\text{AP}$ metrics on COCO 2017. The $Sup.$ column denotes the type of supervision used for training including mask supervision ($\mathcal{M}$), saliency maps ($\mathcal{S}$), image-level labels ($\mathcal{I}$),  and SAM models ($\mathcal{A}$). We mark the best WSIS results in bold.
}
\vspace{-0.08in}
\begin{tabular}{lccccccccccc}
\toprule
                \multicolumn{2}{l}{\multirow{2}{*}{\textbf{Methods}}}& \multicolumn{2}{c}{\multirow{2}{*}{\textbf{Backbone}}} & \multirow{2}{*}{$Sup.$}& \multirow{2}{*}{\textbf{Retrain}} & \multicolumn{3}{c}{\textbf{COCO val 2017}} & \multicolumn{3}{c}{\textbf{COCO test-dev} }\\
 \multicolumn{2}{c}{}&\multicolumn{2}{c}{}&&&$\text{AP}_{\scriptsize50:95}$&$\text{AP}_{\scriptsize50}$&$\text{AP}_{\scriptsize75}$ & $\text{AP}_{\scriptsize50:95}$& $\text{AP}_{\scriptsize50}$&$\text{AP}_{\scriptsize75}$ \\
\toprule
 \multicolumn{8}{l}{\textbf{\textit{Fully-supervised instance segmentation methods.}}} \\
  \multicolumn{2}{l}{Mask R-CNN~\cite{he2017mask}}       & \multicolumn{2}{c}{ResNet-50} &            $\mathcal{M}$& -- & 34.4&      55.1& 36.7 & 33.6 & 55.2&35.3 \\
 \midrule
\multicolumn{8}{l}{\textbf{\textit{WSIS methods with image-level supervision.}}} \\
 \multicolumn{2}{l}{WS-JDS~\cite{shen2019cyclic}}                  & \multicolumn{2}{c}{VGG-16} &        $\mathcal{I}$& -- & 6.1&  11.7& 5.5 & --& --&--\\
                \multicolumn{2}{l}{PDSL~\cite{shen2021parallel}}              & \multicolumn{2}{c}{ResNet18-WS} &        
$\mathcal{I}$& -- & 6.3&      13.1& 5.0 & --& --&--\\
 \multicolumn{2}{l}{Fan et al.~\cite{fan2018wsss}}& \multicolumn{2}{c}{ResNet-101}& $\mathcal{I+S}$& Mask R-CNN & --& --& --& 13.7& 25.5&
13.5 \\
 \multicolumn{2}{l}{LIID~\cite{liu2020liid}}& \multicolumn{2}{c}{ResNet-50}& $\mathcal{I+S}$& Mask R-CNN & --& --& --& 16.0& 27.1&16.5 \\
 \multicolumn{2}{l}{BESTIE~\cite{Kim2022bestie}}& \multicolumn{2}{c}{HRNet-W48}& $\mathcal{I}$& Mask R-CNN & 14.3&28.0& 13.2 & 14.4& 28.0&13.5 \\
 \multicolumn{2}{l}{CIM~\cite{zecheng2023cim}}& \multicolumn{2}{c}{ResNet-50}& $\mathcal{I}$& Mask R-CNN & 17.0& 29.4& 17.0 & 17.2& 29.7&17.3 \\
  \midrule
 \multicolumn{8}{l}{\textbf{\textit{Ours.}}} \\
  \multicolumn{2}{l}{\name{}}& \multicolumn{2}{c}{ResNet-50}& $\mathcal{I + A}$      & Mask R-CNN & 20.6 & 33.9 & 22.0 & 21.0 & 34.5 & 22.2 \\
  \multicolumn{2}{l}{\name{}}& \multicolumn{2}{c}{ResNet-50}& $\mathcal{I + A}$      & Mask2Former & \textbf{25.2} & \textbf{38.4} & \textbf{27.0}& \textbf{25.9} & \textbf{39.9} & \textbf{27.9} \\
 \bottomrule
 
\end{tabular}

\label{tab:perf_wsis_coco}
\vspace{-0.05 in}

\end{table*}

\vspace{-.1in}
\paragraph{Implementation Details}
For \name{} proposals generation, we adopt the WeakTr~\cite{zhu2023weaktr} with DeiT-S~\cite{touvron2021deit} for generating classification clues, the SAM~\cite{kirillov2023sam} with ViT-H~\cite{dosovitskiy2020vit} to generate proposals.
For WSOD pipeline, we use the WSOD networks, i.e., OICR~\cite{tang2017multiple}, and MIST~\cite{ren2020instance}, with the VGG-16~\cite{han2021vgg} to generate pseudo ground truth (PGT), and FSOD networks, i.e., Faster R-CNN~\cite{ren2015faster} and DINO~\cite{zhang2022dino}, with the ResNet-50~\cite{he2016resnet} to retrain. 
As for the WSIS, we use SAM-ViT-H to generate pseudo instance labels and train the R-CNN-based and query-based methods, i.e., Mask R-CNN~\cite{he2017mask} and Mask2former~\cite{cheng2022mask2f}, respectively.
All hyper-parameters in Alg.~\ref{alg:peak_points_extraction} and Alg.~\ref{alg:adaptive_pgt_generation} are following the default manners as ~\citet{zhu2023weaktr} and ~\citet{sui2022sos-wsods}.

\vspace{-.05in}
\subsection{Comparisons with State-of-the-art Methods}

\begin{table*}[htp]
\centering
\small
\caption{Ablation studies for \name{} prompts on PASCAL VOC 2007. We evaluate the average number of proposals, recall, and WSOD performance by MIST~\cite{ren2020instance}.}
\vspace{-0.08in}
\begin{tabular}{ccccc|c|ccc|c}
\toprule
 \multirow{2}{*}{SS} & \multirow{2}{*}{Dense Sample} & \multirow{2}{*}{CAM$_{fine}$} & \multirow{2}{*}{CAM$_{coarse}$} & \multirow{2}{*}{Cross Attn.} & \multirow{2}{*}{Num.} & \multicolumn{3}{c|}{Recall}  & \multirow{2}{*}{$\text{AP}_{\scriptsize{50}}$} \\
 & & & & & & IoU=0.50 & IoU=0.75 & IoU=0.90 & \\
\midrule
 \Checkmark       &&&& & 2001 & 92.6 & 57.7 & 19.2 & 54.9 \\
\midrule
         &\Checkmark       &         &            &            &   129   &    79.6   &    50.7   &   24.3   & 45.2 \\
         &\Checkmark       &    \Checkmark     &            &            &   151   &    88.9   &  67.0     &    37.2 & 63.3\scriptsize\color{red}+18.1 \\
         &\Checkmark       &    \Checkmark     &      \Checkmark      &            &   174   &    90.6   &   70.1    &    40.1  & 65.5\scriptsize\color{red}+20.3 \\
         &\Checkmark       &    \Checkmark     &      \Checkmark      &      \Checkmark      &   213   &    95.6   &   75.0    &   42.1  & 67.4\scriptsize\color{red}+22.2 \\

\bottomrule
\end{tabular}

\vspace{-0.05in}
\label{tab:recall}
\end{table*}

\paragraph{Weakly-supervised object detection}

We present the quantitative WSOD results in Table.~\ref{tab:perf_Wsod}.
Compared with our WSOD baseline methods, i.e., OICR and MIST, the proposed \name{} achieves over 10\% improvements on all metrics. The results of \name{} (MIST) surpass all WSOD methods on all metrics, which demonstrate the effectiveness of \name{} proposals.
Compared with WSOD methods retrained by pseudo ground truth (PGT), the \name{} (MIST) with Faster R-CNN retraining still outperforms the SoS-WSOD~\cite{sui2022sos-wsods} and W2N~\cite{huang2022w2nswitching} on all metrics, and the \name{} (MIST) with DINO retraining even has comparable performance with fully-supervised Faster R-CNN. The retraining results demonstrate the effectiveness of the proposed WSOD pipeline, which includes the adaptive PGT generation and RoI drop retraining.
Compared with concurrent work, WSOVOD~\cite{lin2023weakly}, which also incorporates SAM, our \name{} (MIST) also achieves better performance.

\vspace{-.08in}
\paragraph{Weakly-supervised instance segmentation}

We first present the quantitative WSIS results of the PASCAL VOC 2012 $val$ set in Table~\ref{tab:perf_wsis_voc}. The proposed \name{} with Mask R-CNN retraining achieves the best performance, which demonstrates the \name{} can benefit WSIS effectively. Furthermore, the pseudo instance labels generated by \name{} can also be used by the modern query-based methods, e.g., Mask2Former~\cite{cheng2022mask2f}, which achieves the best results.

We then show the quantitative WSIS results on COCO 2017 $val$ and $test$ sets. On these more challenging benchmarks, \name{} with Mask R-CNN retraining achieves better results than CIM~\cite{zecheng2023cim}. Besides, \name{} with Mask2Former also presents the best results.

\vspace{-.1in}

\begin{table}[htp]
\small
\centering
\caption{Ablation studies for adaptive PGT generation and RoI drop regularization. We present the results on the PASCAL VOC 2007 $test$ set.}
\vspace{-.12in}
\begin{subtable}{1.0\linewidth}
\centering
\caption{Ablation studies for the anchor-based detector, i.e., Faster R-CNN~\cite{ren2015faster}.}
\begin{tabular}{ccc|c}
\toprule
Top-1 PGT & Adaptive PGT & RoI Drop & \text{AP}$_{50}$ \\
\midrule
 \Checkmark       &&& 68.4 \\
\midrule
         &\Checkmark       &         &     70.7\scriptsize\color{red}+2.3         \\
         &\Checkmark       &    \Checkmark     &  71.8\scriptsize\color{red}+3.4  \\

\bottomrule
\end{tabular}
\label{tab: abla_frrcnn}

\centering
\caption{Ablation studies for the query-based detector, i.e., DINO~\cite{zhang2022dino}.}
\begin{tabular}{ccc|c}
\toprule
Top-1 PGT & Adaptive PGT & Query Drop & \text{AP}$_{50}$ \\
\midrule
 \Checkmark       &&& 71.1 \\
\midrule
         &\Checkmark       &         &     72.8\scriptsize\color{red}+1.7         \\
         &\Checkmark       &    \Checkmark     & 73.4\scriptsize\color{red}+2.3  \\

\bottomrule
\end{tabular}
\label{tab: abla_dino}

\vspace{-0.1in}
\end{subtable}
\label{tab:retrain_voc_components}
\vspace{-0.1in}
\end{table}

\subsection{Ablation Studies}
In this section, we present the ablation studies to evaluate the improvements brought by the proposed methods, i.e., \name{} prompts, adaptive PGT generation, and RoI drop retraining. 

Due to the limitation of pages, we leave more ablation studies in the supplementary material, including additional efficiency analysis, sensitivity analysis, qualitative analysis, discussions, etc.

\vspace{-0.08 in}

\begin{table}[htp]
\small
\centering
\caption{Efficiency comparison between Selective Search and our \name{} during the training on the PASCAL VOC 2007. `Num.' is the number of proposals, `T$_\text{Proposals}$' is the time consumption for generating proposals, `T$_\text{WSOD}$' is the time consumption for training the WSOD network, i.e., MIST~\cite{ren2020instance}, and `M$_\text{WSOD}$' is the GPU memory cost for each GPU card. }
\vspace{-0.08in}
\begin{tabular}{lcccc}
\toprule
 & Num. & T$_\text{Proposals}$ & T$_\text{WSOD}$ & M$_\text{WSOD}$ \\
\midrule
 SS~\cite{Uijlings2013SelectiveSF}       & 2001 & 11.6 hrs & 16 hrs & 17810 MiB \\ %
\midrule
 Ours         & 213\scriptsize\color{red}-89.4\% & 4 hrs\scriptsize\color{red}-65.5\% & 9 hrs\scriptsize\color{red}-43.8\% & 5667 MiB\scriptsize\color{red}-68.2\% \\
\bottomrule
\end{tabular}
\label{tab: comp_efficiency}
\vspace{-.15in}
\end{table}

\begin{figure*}[htp]

    \centering
    \includegraphics[width=1.0\linewidth]{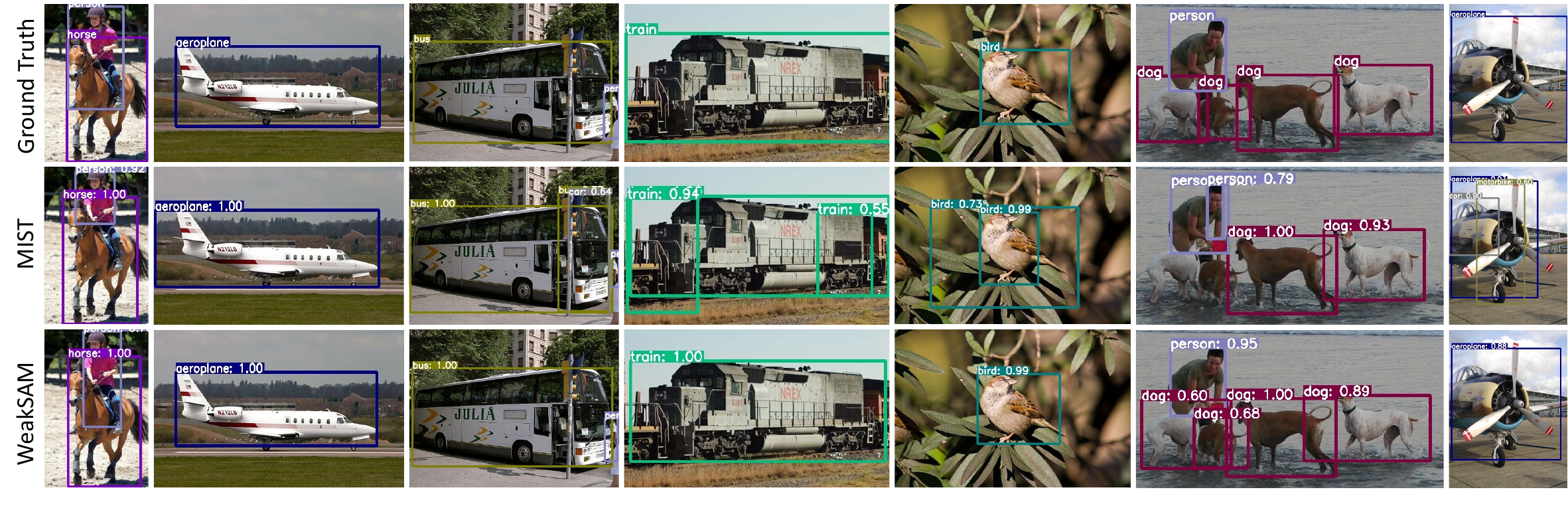}
    \vspace{-.4 in}
    \caption{Visualization of the weakly-supervised object detection on the PASCAL VOC 2007 $test$ set.}
    \label{fig: vis_comp_det}
\vspace{-.1in}
\end{figure*}

\begin{figure*}[htp]

    \centering
    \includegraphics[width=1.0\linewidth]{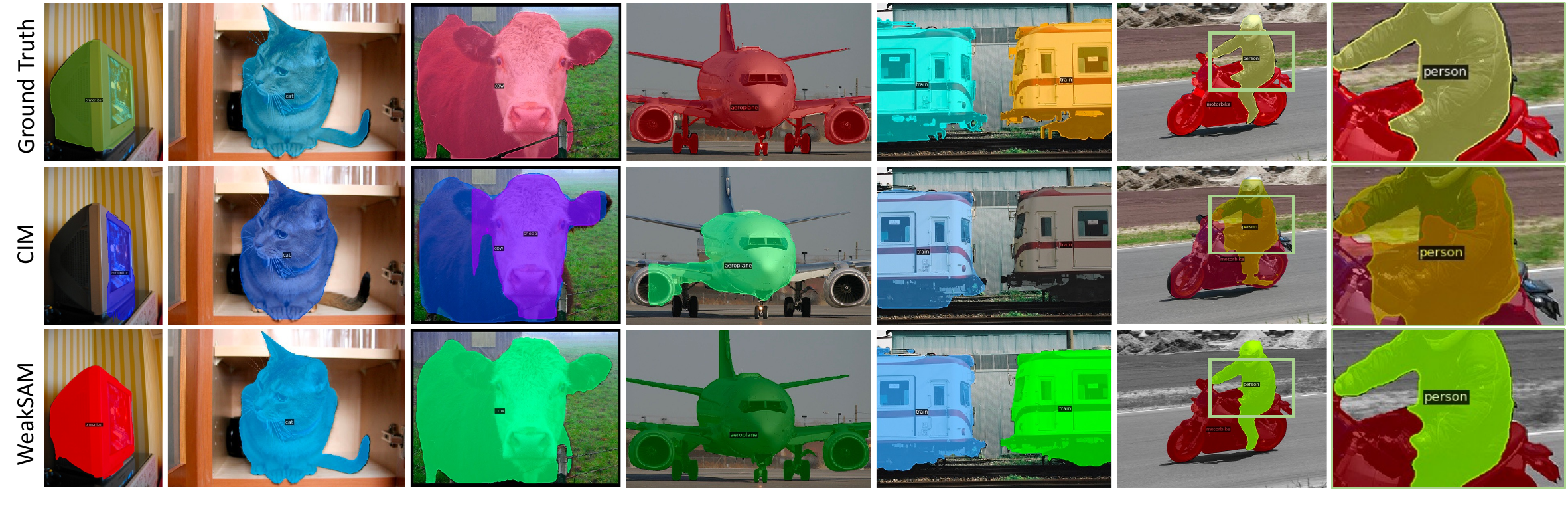}
    \vspace{-.4 in}
    \caption{Visualization of the weakly-supervised instance segmentation on the PASCAL VOC 2012 $val$ set.}
    \label{fig: vis_comp_ins}
\vspace{-.1in}
\end{figure*}

\paragraph{Improvements of \name{} Prompts}

To further analyze the improvements brought by the proposed \name{} prompts, we conduct ablation experiments for different prompts as shown in Table~\ref{tab:recall}. Here, we use the Selective Search~\cite{Uijlings2013SelectiveSF} as the baseline method and list the proposals' number, recall, and corresponding WSOD performance. When only using the densely sampled points as SAM prompts, the generated proposals can achieve 5.1\% higher Recall (IoU=0.90), and 9.7\% lower \text{AP$_{50}$} for MIST. After adding peak CAM points and peak cross attention points as prompts, we can achieve higher recall and \text{AP$_{50}$} through only 213 proposals on average. The results demonstrate the effectiveness of \name{} prompts.

\vspace{-0.1 in}

\paragraph{Improvements of WSOD Pipeline}

To further analyze the improvements brought by the proposed \name{} WSOD pipeline, we conduct ablation experiments for adaptive PGT generation and RoI drop regularization in Table~\ref{tab:retrain_voc_components}. 
Here, we follow the common practice to set a baseline that uses the predicted boxes with the top-1 score as PGT and plain Faster R-CNN as the retraining network. 
The results show that both adaptive PGT generation and RoI drop regularization can help improve the \text{AP$_{50}$} of the detector.
Furthermore, both the RoI-based detector, Faster R-CNN~\cite{ren2015faster} and query-based detecotr, DINO~\cite{zhang2022dino}, can benefit from the proposed WSOD techniques.

\vspace{-.1in}

\subsection{Efficiency Comparison}
To further analyze the efficiency improvement brought by our \name{}, we present the efficiency comparison between Selective Search~\cite{Uijlings2013SelectiveSF} and our \name{} on a machine with 4 GPU cards, as shown in Table~\ref{tab: comp_efficiency}. Our \name{} reduces the number of proposals by 89.4\%, the proposal generation time by 65.5\%, the WSOD network training time by 43.8\%, and the GPU memory cost by 68.2\%. The results demonstrate the significant efficiency improvement brought by the proposed \name{}.

\subsection{Visualization Results}

Fig.\ref{fig: vis_comp_det} presents the object detection results using \name{} (MIST), showing its capability to accurately capture entire objects without generating excessive noisy bounding boxes. In Fig.\ref{fig: vis_comp_ins}, the instance segmentation results of \name{} Mask2Former retraining are showcased. The results indicate effective segmentation of entire instances with a notable reduction in overlapping segments.

\section{Conclusion}
\label{sec:conclu}
In this paper, we introduce \name{}, a novel framework utilizing the Segment Anything Model (SAM) for weakly-supervised instance-level recognition, demonstrating leading performance in WSOD and WSIS benchmarks. Different from the original SAM, which requires interaction and can not be aware of categories, \name{} represents an innovative fusion of SAM with weakly-supervised learning (WSL), overcoming the redundancy problem of WSOD proposals. To further address WSOD issues such as pseudo ground truth (PGT) incompleteness and noisy PGT instances, our approach includes adaptive PGT generation and a Region of Interest (RoI) drop regularization. The adaptability of \name{} is further showcased through its extension to weakly-supervised instance segmentation (WSIS). Our work aims to inspire further research with SAM and WSL, contributing significantly to the development of a universal framework for weakly-supervised recognition.

\section*{Acknowledgement}
This work was supported by National Natural Science Foundation of China (NSFC) under Grant No. 62276108.

\bibliographystyle{ACM-Reference-Format}
\balance
\bibliography{weaksam}


\begin{thebibliography}{104}


\ifx \showCODEN    \undefined \def \showCODEN     #1{\unskip}     \fi
\ifx \showDOI      \undefined \def \showDOI       #1{#1}\fi
\ifx \showISBNx    \undefined \def \showISBNx     #1{\unskip}     \fi
\ifx \showISBNxiii \undefined \def \showISBNxiii  #1{\unskip}     \fi
\ifx \showISSN     \undefined \def \showISSN      #1{\unskip}     \fi
\ifx \showLCCN     \undefined \def \showLCCN      #1{\unskip}     \fi
\ifx \shownote     \undefined \def \shownote      #1{#1}          \fi
\ifx \showarticletitle \undefined \def \showarticletitle #1{#1}   \fi
\ifx \showURL      \undefined \def \showURL       {\relax}        \fi
\providecommand\bibfield[2]{#2}
\providecommand\bibinfo[2]{#2}
\providecommand\natexlab[1]{#1}
\providecommand\showeprint[2][]{arXiv:#2}

\bibitem[Ahn et~al\mbox{.}(2019)]%
        {Ahn2019irn}
\bibfield{author}{\bibinfo{person}{Jiwoon Ahn}, \bibinfo{person}{Sunghyun Cho}, {and} \bibinfo{person}{Suha Kwak}.} \bibinfo{year}{2019}\natexlab{}.
\newblock \showarticletitle{Weakly Supervised Learning of Instance Segmentation With Inter-Pixel Relations}. In \bibinfo{booktitle}{\emph{Proc. of CVPR}}.
\newblock


\bibitem[Arun et~al\mbox{.}(2020)]%
        {arun2020weakly}
\bibfield{author}{\bibinfo{person}{Aditya Arun}, \bibinfo{person}{CV Jawahar}, {and} \bibinfo{person}{M~Pawan Kumar}.} \bibinfo{year}{2020}\natexlab{}.
\newblock \showarticletitle{Weakly supervised instance segmentation by learning annotation consistent instances}. In \bibinfo{booktitle}{\emph{Proc. of ECCV}}.
\newblock


\bibitem[Arun et~al\mbox{.}(2019)]%
        {Arun_2019_CVPR}
\bibfield{author}{\bibinfo{person}{Aditya Arun}, \bibinfo{person}{C.~V. Jawahar}, {and} \bibinfo{person}{M.~Pawan Kumar}.} \bibinfo{year}{2019}\natexlab{}.
\newblock \showarticletitle{Dissimilarity Coefficient Based Weakly Supervised Object Detection}. In \bibinfo{booktitle}{\emph{Proc. of CVPR}}.
\newblock


\bibitem[Bilen and Vedaldi(2016)]%
        {bilen2016wsddn}
\bibfield{author}{\bibinfo{person}{Hakan Bilen} {and} \bibinfo{person}{Andrea Vedaldi}.} \bibinfo{year}{2016}\natexlab{}.
\newblock \showarticletitle{Weakly Supervised Deep Detection Networks}. In \bibinfo{booktitle}{\emph{Proc. of CVPR}}.
\newblock


\bibitem[Chen et~al\mbox{.}(2021b)]%
        {mmChen2021wsss}
\bibfield{author}{\bibinfo{person}{Jianjun Chen}, \bibinfo{person}{Shancheng Fang}, \bibinfo{person}{Hongtao Xie}, \bibinfo{person}{Zheng-Jun Zha}, \bibinfo{person}{Yue Hu}, {and} \bibinfo{person}{Jianlong Tan}.} \bibinfo{year}{2021}\natexlab{b}.
\newblock \showarticletitle{End-to-end Boundary Exploration for Weakly-supervised Semantic Segmentation}. In \bibinfo{booktitle}{\emph{ACM MM}}. \bibinfo{numpages}{10}~pages.
\newblock
\showISBNx{9781450386517}


\bibitem[Chen et~al\mbox{.}(2022)]%
        {chen2022point}
\bibfield{author}{\bibinfo{person}{Pengfei Chen}, \bibinfo{person}{Xuehui Yu}, \bibinfo{person}{Xumeng Han}, \bibinfo{person}{Najmul Hassan}, \bibinfo{person}{Kai Wang}, \bibinfo{person}{Jiachen Li}, \bibinfo{person}{Jian Zhao}, \bibinfo{person}{Humphrey Shi}, \bibinfo{person}{Zhenjun Han}, {and} \bibinfo{person}{Qixiang Ye}.} \bibinfo{year}{2022}\natexlab{}.
\newblock \showarticletitle{Point-to-box network for accurate object detection via single point supervision}. In \bibinfo{booktitle}{\emph{Proc. of ECCV}}.
\newblock


\bibitem[Chen et~al\mbox{.}(2023)]%
        {chen2023sepl}
\bibfield{author}{\bibinfo{person}{Tianle Chen}, \bibinfo{person}{Zheda Mai}, \bibinfo{person}{Ruiwen Li}, {and} \bibinfo{person}{Wei-lun Chao}.} \bibinfo{year}{2023}\natexlab{}.
\newblock \showarticletitle{Segment anything model (sam) enhanced pseudo labels for weakly supervised semantic segmentation}.
\newblock \bibinfo{journal}{\emph{ArXiv preprint}} (\bibinfo{year}{2023}).
\newblock


\bibitem[Chen et~al\mbox{.}(2021a)]%
        {mmChen2021wsol}
\bibfield{author}{\bibinfo{person}{Zhiwei Chen}, \bibinfo{person}{Liujuan Cao}, \bibinfo{person}{Yunhang Shen}, \bibinfo{person}{Feihong Lian}, \bibinfo{person}{Yongjian Wu}, {and} \bibinfo{person}{Rongrong Ji}.} \bibinfo{year}{2021}\natexlab{a}.
\newblock \showarticletitle{E2Net: Excitative-Expansile Learning for Weakly Supervised Object Localization}. In \bibinfo{booktitle}{\emph{ACM MM}}. \bibinfo{numpages}{9}~pages.
\newblock
\showISBNx{9781450386517}


\bibitem[Chen et~al\mbox{.}(2020)]%
        {Chen_2020_CVPR}
\bibfield{author}{\bibinfo{person}{Ze Chen}, \bibinfo{person}{Zhihang Fu}, \bibinfo{person}{Rongxin Jiang}, \bibinfo{person}{Yaowu Chen}, {and} \bibinfo{person}{Xian{-}Sheng Hua}.} \bibinfo{year}{2020}\natexlab{}.
\newblock \showarticletitle{{SLV:} Spatial Likelihood Voting for Weakly Supervised Object Detection}. In \bibinfo{booktitle}{\emph{Proc. of CVPR}}.
\newblock


\bibitem[Cheng et~al\mbox{.}(2022)]%
        {cheng2022mask2f}
\bibfield{author}{\bibinfo{person}{Bowen Cheng}, \bibinfo{person}{Ishan Misra}, \bibinfo{person}{Alexander~G. Schwing}, \bibinfo{person}{Alexander Kirillov}, {and} \bibinfo{person}{Rohit Girdhar}.} \bibinfo{year}{2022}\natexlab{}.
\newblock \showarticletitle{Masked-attention Mask Transformer for Universal Image Segmentation}. In \bibinfo{booktitle}{\emph{Proc. of CVPR}}.
\newblock


\bibitem[Cheng et~al\mbox{.}(2023)]%
        {cheng2023boxteacher}
\bibfield{author}{\bibinfo{person}{Tianheng Cheng}, \bibinfo{person}{Xinggang Wang}, \bibinfo{person}{Shaoyu Chen}, \bibinfo{person}{Qian Zhang}, {and} \bibinfo{person}{Wenyu Liu}.} \bibinfo{year}{2023}\natexlab{}.
\newblock \showarticletitle{Boxteacher: Exploring high-quality pseudo labels for weakly supervised instance segmentation}. In \bibinfo{booktitle}{\emph{Proc. of CVPR}}.
\newblock


\bibitem[Diba et~al\mbox{.}(2017)]%
        {diba2017weakly}
\bibfield{author}{\bibinfo{person}{Ali Diba}, \bibinfo{person}{Vivek Sharma}, \bibinfo{person}{Ali~Mohammad Pazandeh}, \bibinfo{person}{Hamed Pirsiavash}, {and} \bibinfo{person}{Luc~Van Gool}.} \bibinfo{year}{2017}\natexlab{}.
\newblock \showarticletitle{Weakly Supervised Cascaded Convolutional Networks}. In \bibinfo{booktitle}{\emph{Proc. of CVPR}}.
\newblock


\bibitem[Dosovitskiy et~al\mbox{.}(2021)]%
        {dosovitskiy2020vit}
\bibfield{author}{\bibinfo{person}{Alexey Dosovitskiy}, \bibinfo{person}{Lucas Beyer}, \bibinfo{person}{Alexander Kolesnikov}, \bibinfo{person}{Dirk Weissenborn}, \bibinfo{person}{Xiaohua Zhai}, \bibinfo{person}{Thomas Unterthiner}, \bibinfo{person}{Mostafa Dehghani}, \bibinfo{person}{Matthias Minderer}, \bibinfo{person}{Georg Heigold}, \bibinfo{person}{Sylvain Gelly}, \bibinfo{person}{Jakob Uszkoreit}, {and} \bibinfo{person}{Neil Houlsby}.} \bibinfo{year}{2021}\natexlab{}.
\newblock \showarticletitle{An Image is Worth 16x16 Words: Transformers for Image Recognition at Scale}. In \bibinfo{booktitle}{\emph{Proc. of ICLR}}.
\newblock


\bibitem[Duan et~al\mbox{.}(2022)]%
        {duan2022rda}
\bibfield{author}{\bibinfo{person}{Yue Duan}, \bibinfo{person}{Lei Qi}, \bibinfo{person}{Lei Wang}, \bibinfo{person}{Luping Zhou}, {and} \bibinfo{person}{Yinghuan Shi}.} \bibinfo{year}{2022}\natexlab{}.
\newblock \showarticletitle{Rda: Reciprocal distribution alignment for robust semi-supervised learning}. In \bibinfo{booktitle}{\emph{Proc. of ECCV}}.
\newblock


\bibitem[Duan et~al\mbox{.}(2024a)]%
        {duan2024mutexmatch}
\bibfield{author}{\bibinfo{person}{Yue Duan}, \bibinfo{person}{Zhen Zhao}, \bibinfo{person}{Lei Qi}, \bibinfo{person}{Lei Wang}, \bibinfo{person}{Luping Zhou}, \bibinfo{person}{Yinghuan Shi}, {and} \bibinfo{person}{Yang Gao}.} \bibinfo{year}{2024}\natexlab{a}.
\newblock \showarticletitle{MutexMatch: Semi-Supervised Learning With Mutex-Based Consistency Regularization}.
\newblock \bibinfo{journal}{\emph{TNNLS}} (\bibinfo{year}{2024}).
\newblock


\bibitem[Duan et~al\mbox{.}(2023)]%
        {duan2023towards}
\bibfield{author}{\bibinfo{person}{Yue Duan}, \bibinfo{person}{Zhen Zhao}, \bibinfo{person}{Lei Qi}, \bibinfo{person}{Luping Zhou}, \bibinfo{person}{Lei Wang}, {and} \bibinfo{person}{Yinghuan Shi}.} \bibinfo{year}{2023}\natexlab{}.
\newblock \showarticletitle{Towards semi-supervised learning with non-random missing labels}. In \bibinfo{booktitle}{\emph{Proc. of ICCV}}.
\newblock


\bibitem[Duan et~al\mbox{.}(2024b)]%
        {duan2024roll}
\bibfield{author}{\bibinfo{person}{Yue Duan}, \bibinfo{person}{Zhen Zhao}, \bibinfo{person}{Lei Qi}, \bibinfo{person}{Luping Zhou}, \bibinfo{person}{Lei Wang}, {and} \bibinfo{person}{Yinghuan Shi}.} \bibinfo{year}{2024}\natexlab{b}.
\newblock \showarticletitle{Roll with the Punches: Expansion and Shrinkage of Soft Label Selection for Semi-supervised Fine-Grained Learning}. In \bibinfo{booktitle}{\emph{Proc. of AAAI}}.
\newblock


\bibitem[Everingham et~al\mbox{.}(2015)]%
        {everingham2015voc}
\bibfield{author}{\bibinfo{person}{Mark Everingham}, \bibinfo{person}{SM~Ali Eslami}, \bibinfo{person}{Luc Van~Gool}, \bibinfo{person}{Christopher~KI Williams}, \bibinfo{person}{John Winn}, {and} \bibinfo{person}{Andrew Zisserman}.} \bibinfo{year}{2015}\natexlab{}.
\newblock \showarticletitle{The pascal visual object classes challenge: A retrospective}.
\newblock \bibinfo{journal}{\emph{IJCV}} (\bibinfo{year}{2015}).
\newblock


\bibitem[Fan et~al\mbox{.}(2018)]%
        {fan2018wsss}
\bibfield{author}{\bibinfo{person}{Ruochen Fan}, \bibinfo{person}{Qibin Hou}, \bibinfo{person}{Ming-Ming Cheng}, \bibinfo{person}{Gang Yu}, \bibinfo{person}{Ralph~R Martin}, {and} \bibinfo{person}{Shi-Min Hu}.} \bibinfo{year}{2018}\natexlab{}.
\newblock \showarticletitle{Associating inter-image salient instances for weakly supervised semantic segmentation}. In \bibinfo{booktitle}{\emph{Proc. of ECCV}}.
\newblock


\bibitem[Fu et~al\mbox{.}(2020)]%
        {fu2020wstriplet}
\bibfield{author}{\bibinfo{person}{Daniel~Y. Fu}, \bibinfo{person}{Mayee~F. Chen}, \bibinfo{person}{Frederic Sala}, \bibinfo{person}{Sarah~M. Hooper}, \bibinfo{person}{Kayvon Fatahalian}, {and} \bibinfo{person}{Christopher R{\'{e}}}.} \bibinfo{year}{2020}\natexlab{}.
\newblock \showarticletitle{Fast and Three-rious: Speeding Up Weak Supervision with Triplet Methods}. In \bibinfo{booktitle}{\emph{Proc. of ICML}}.
\newblock


\bibitem[Gao et~al\mbox{.}(2018)]%
        {gao2018c}
\bibfield{author}{\bibinfo{person}{Mingfei Gao}, \bibinfo{person}{Ang Li}, \bibinfo{person}{Ruichi Yu}, \bibinfo{person}{Vlad~I Morariu}, {and} \bibinfo{person}{Larry~S Davis}.} \bibinfo{year}{2018}\natexlab{}.
\newblock \showarticletitle{C-wsl: Count-guided weakly supervised localization}. In \bibinfo{booktitle}{\emph{Proc. of ECCV}}.
\newblock


\bibitem[Ge et~al\mbox{.}(2019)]%
        {ge2019label-penet}
\bibfield{author}{\bibinfo{person}{Weifeng Ge}, \bibinfo{person}{Weilin Huang}, \bibinfo{person}{Sheng Guo}, {and} \bibinfo{person}{Matthew~R. Scott}.} \bibinfo{year}{2019}\natexlab{}.
\newblock \showarticletitle{Label-PEnet: Sequential Label Propagation and Enhancement Networks for Weakly Supervised Instance Segmentation}. In \bibinfo{booktitle}{\emph{Proc. of ICCV}}.
\newblock


\bibitem[Han et~al\mbox{.}(2021)]%
        {han2021vgg}
\bibfield{author}{\bibinfo{person}{Kai Han}, \bibinfo{person}{An Xiao}, \bibinfo{person}{Enhua Wu}, \bibinfo{person}{Jianyuan Guo}, \bibinfo{person}{Chunjing Xu}, {and} \bibinfo{person}{Yunhe Wang}.} \bibinfo{year}{2021}\natexlab{}.
\newblock \showarticletitle{Transformer in Transformer}. In \bibinfo{booktitle}{\emph{Proc. of NeurIPS}}.
\newblock


\bibitem[He et~al\mbox{.}(2017)]%
        {he2017mask}
\bibfield{author}{\bibinfo{person}{Kaiming He}, \bibinfo{person}{Georgia Gkioxari}, \bibinfo{person}{Piotr Doll{\'{a}}r}, {and} \bibinfo{person}{Ross~B. Girshick}.} \bibinfo{year}{2017}\natexlab{}.
\newblock \showarticletitle{Mask {R-CNN}}. In \bibinfo{booktitle}{\emph{Proc. of ICCV}}.
\newblock


\bibitem[He et~al\mbox{.}(2016)]%
        {he2016resnet}
\bibfield{author}{\bibinfo{person}{Kaiming He}, \bibinfo{person}{Xiangyu Zhang}, \bibinfo{person}{Shaoqing Ren}, {and} \bibinfo{person}{Jian Sun}.} \bibinfo{year}{2016}\natexlab{}.
\newblock \showarticletitle{Deep Residual Learning for Image Recognition}. In \bibinfo{booktitle}{\emph{Proc. of CVPR}}.
\newblock


\bibitem[Hsieh et~al\mbox{.}(2023)]%
        {Hsieh_2023_ICCV}
\bibfield{author}{\bibinfo{person}{Yu-Hsing Hsieh}, \bibinfo{person}{Guan-Sheng Chen}, \bibinfo{person}{Shun-Xian Cai}, \bibinfo{person}{Ting-Yun Wei}, \bibinfo{person}{Huei-Fang Yang}, {and} \bibinfo{person}{Chu-Song Chen}.} \bibinfo{year}{2023}\natexlab{}.
\newblock \showarticletitle{Class-incremental Continual Learning for Instance Segmentation with Image-level Weak Supervision}. In \bibinfo{booktitle}{\emph{ICCV}}.
\newblock


\bibitem[Hsu et~al\mbox{.}(2019)]%
        {hsu2019bbtp}
\bibfield{author}{\bibinfo{person}{Cheng{-}Chun Hsu}, \bibinfo{person}{Kuang{-}Jui Hsu}, \bibinfo{person}{Chung{-}Chi Tsai}, \bibinfo{person}{Yen{-}Yu Lin}, {and} \bibinfo{person}{Yung{-}Yu Chuang}.} \bibinfo{year}{2019}\natexlab{}.
\newblock \showarticletitle{Weakly Supervised Instance Segmentation using the Bounding Box Tightness Prior}. In \bibinfo{booktitle}{\emph{Proc. of NeurIPS}}.
\newblock


\bibitem[Hu et~al\mbox{.}(2020)]%
        {hu2020weakly}
\bibfield{author}{\bibinfo{person}{Zheng Hu}, \bibinfo{person}{Zhi Liu}, \bibinfo{person}{Gongyang Li}, \bibinfo{person}{Linwei Ye}, \bibinfo{person}{Lei Zhou}, {and} \bibinfo{person}{Yang Wang}.} \bibinfo{year}{2020}\natexlab{}.
\newblock \showarticletitle{Weakly supervised instance segmentation using multi-stage erasing refinement and saliency-guided proposals ordering}.
\newblock \bibinfo{journal}{\emph{JVCI}} (\bibinfo{year}{2020}).
\newblock


\bibitem[Huang et~al\mbox{.}(2022)]%
        {huang2022w2nswitching}
\bibfield{author}{\bibinfo{person}{Zitong Huang}, \bibinfo{person}{Yiping Bao}, \bibinfo{person}{Bowen Dong}, \bibinfo{person}{Erjin Zhou}, {and} \bibinfo{person}{Wangmeng Zuo}.} \bibinfo{year}{2022}\natexlab{}.
\newblock \bibinfo{title}{W2N:Switching From Weak Supervision to Noisy Supervision for Object Detection}.
\newblock
\newblock
\showeprint[arxiv]{2207.12104}~[cs.CV]


\bibitem[Huang et~al\mbox{.}(2020)]%
        {huang2020comprehensive}
\bibfield{author}{\bibinfo{person}{Zeyi Huang}, \bibinfo{person}{Yang Zou}, \bibinfo{person}{B.~V. K.~Vijaya Kumar}, {and} \bibinfo{person}{Dong Huang}.} \bibinfo{year}{2020}\natexlab{}.
\newblock \showarticletitle{Comprehensive Attention Self-Distillation for Weakly-Supervised Object Detection}. In \bibinfo{booktitle}{\emph{Proc. of NeurIPS}}.
\newblock


\bibitem[Hwang et~al\mbox{.}(2021)]%
        {Hwang_2021_WACV}
\bibfield{author}{\bibinfo{person}{Jaedong Hwang}, \bibinfo{person}{Seohyun Kim}, \bibinfo{person}{Jeany Son}, {and} \bibinfo{person}{Bohyung Han}.} \bibinfo{year}{2021}\natexlab{}.
\newblock \showarticletitle{Weakly Supervised Instance Segmentation by Deep Community Learning}. In \bibinfo{booktitle}{\emph{Proc. of WACV}}.
\newblock


\bibitem[Jia et~al\mbox{.}(2021)]%
        {jia2021gradingnet}
\bibfield{author}{\bibinfo{person}{Qifei Jia}, \bibinfo{person}{Shikui Wei}, \bibinfo{person}{Tao Ruan}, \bibinfo{person}{Yufeng Zhao}, {and} \bibinfo{person}{Yao Zhao}.} \bibinfo{year}{2021}\natexlab{}.
\newblock \showarticletitle{GradingNet: Towards Providing Reliable Supervisions for Weakly Supervised Object Detection by Grading the Box Candidates}. In \bibinfo{booktitle}{\emph{Proc. of AAAI}}.
\newblock


\bibitem[Jiang and Yang(2023)]%
        {jiang2023sam-pseudo-labeler}
\bibfield{author}{\bibinfo{person}{Peng-Tao Jiang} {and} \bibinfo{person}{Yuqi Yang}.} \bibinfo{year}{2023}\natexlab{}.
\newblock \showarticletitle{Segment Anything is A Good Pseudo-label Generator for Weakly Supervised Semantic Segmentation}.
\newblock \bibinfo{journal}{\emph{ArXiv preprint}} (\bibinfo{year}{2023}).
\newblock


\bibitem[Jie et~al\mbox{.}(2017)]%
        {jie2017self-taught}
\bibfield{author}{\bibinfo{person}{Zequn Jie}, \bibinfo{person}{Yunchao Wei}, \bibinfo{person}{Xiaojie Jin}, \bibinfo{person}{Jiashi Feng}, {and} \bibinfo{person}{Wei Liu}.} \bibinfo{year}{2017}\natexlab{}.
\newblock \showarticletitle{Deep Self-Taught Learning for Weakly Supervised Object Localization}. In \bibinfo{booktitle}{\emph{Proc. of CVPR}}.
\newblock


\bibitem[Khoreva et~al\mbox{.}(2017)]%
        {Khoreva_2017_CVPR}
\bibfield{author}{\bibinfo{person}{Anna Khoreva}, \bibinfo{person}{Rodrigo Benenson}, \bibinfo{person}{Jan~Hendrik Hosang}, \bibinfo{person}{Matthias Hein}, {and} \bibinfo{person}{Bernt Schiele}.} \bibinfo{year}{2017}\natexlab{}.
\newblock \showarticletitle{Simple Does It: Weakly Supervised Instance and Semantic Segmentation}. In \bibinfo{booktitle}{\emph{Proc. of CVPR}}.
\newblock


\bibitem[Kim et~al\mbox{.}(2022)]%
        {Kim2022bestie}
\bibfield{author}{\bibinfo{person}{Beomyoung Kim}, \bibinfo{person}{Youngjoon Yoo}, \bibinfo{person}{Chaeeun Rhee}, {and} \bibinfo{person}{Junmo Kim}.} \bibinfo{year}{2022}\natexlab{}.
\newblock \showarticletitle{Beyond Semantic to Instance Segmentation: Weakly-Supervised Instance Segmentation via Semantic Knowledge Transfer and Self-Refinement}. In \bibinfo{booktitle}{\emph{Proc. of CVPR}}.
\newblock


\bibitem[Kirillov et~al\mbox{.}(2023)]%
        {kirillov2023sam}
\bibfield{author}{\bibinfo{person}{Alexander Kirillov}, \bibinfo{person}{Eric Mintun}, \bibinfo{person}{Nikhila Ravi}, \bibinfo{person}{Hanzi Mao}, \bibinfo{person}{Chloe Rolland}, \bibinfo{person}{Laura Gustafson}, \bibinfo{person}{Tete Xiao}, \bibinfo{person}{Spencer Whitehead}, \bibinfo{person}{Alexander~C. Berg}, \bibinfo{person}{Wan-Yen Lo}, \bibinfo{person}{Piotr Doll{\'a}r}, {and} \bibinfo{person}{Ross Girshick}.} \bibinfo{year}{2023}\natexlab{}.
\newblock \showarticletitle{Segment Anything}.
\newblock \bibinfo{journal}{\emph{ArXiv preprint}} (\bibinfo{year}{2023}).
\newblock


\bibitem[Laptev et~al\mbox{.}({[n.\,d.]})]%
        {laptevcontextlocnet}
\bibfield{author}{\bibinfo{person}{Ivan Laptev}, \bibinfo{person}{Vadim Kantorov}, \bibinfo{person}{Maxime Oquab}, {and} \bibinfo{person}{Minsu Cho}.} \bibinfo{year}{[n.\,d.]}\natexlab{}.
\newblock \showarticletitle{ContextLocNet: Context-aware deep network models for weakly supervised localization}.
\newblock  (\bibinfo{year}{[n.\,d.]}).
\newblock


\bibitem[Laradji et~al\mbox{.}(2019)]%
        {laradji2019masks}
\bibfield{author}{\bibinfo{person}{Issam~H. Laradji}, \bibinfo{person}{David V{\'{a}}zquez}, {and} \bibinfo{person}{Mark Schmidt}.} \bibinfo{year}{2019}\natexlab{}.
\newblock \showarticletitle{Where are the Masks: Instance Segmentation with Image-level Supervision}. In \bibinfo{booktitle}{\emph{Proc. of BMVC}}.
\newblock


\bibitem[Lee et~al\mbox{.}(2021)]%
        {Lee_2021_CVPR}
\bibfield{author}{\bibinfo{person}{Jungbeom Lee}, \bibinfo{person}{Jihun Yi}, \bibinfo{person}{Chaehun Shin}, {and} \bibinfo{person}{Sungroh Yoon}.} \bibinfo{year}{2021}\natexlab{}.
\newblock \showarticletitle{{BBAM:} Bounding Box Attribution Map for Weakly Supervised Semantic and Instance Segmentation}. In \bibinfo{booktitle}{\emph{Proc. of CVPR}}.
\newblock


\bibitem[Li et~al\mbox{.}(2016)]%
        {li2016pda}
\bibfield{author}{\bibinfo{person}{Dong Li}, \bibinfo{person}{Jia{-}Bin Huang}, \bibinfo{person}{Yali Li}, \bibinfo{person}{Shengjin Wang}, {and} \bibinfo{person}{Ming{-}Hsuan Yang}.} \bibinfo{year}{2016}\natexlab{}.
\newblock \showarticletitle{Weakly Supervised Object Localization with Progressive Domain Adaptation}. In \bibinfo{booktitle}{\emph{Proc. of CVPR}}.
\newblock


\bibitem[Li et~al\mbox{.}(2022)]%
        {li2022box}
\bibfield{author}{\bibinfo{person}{Wentong Li}, \bibinfo{person}{Wenyu Liu}, \bibinfo{person}{Jianke Zhu}, \bibinfo{person}{Miaomiao Cui}, \bibinfo{person}{Xian-Sheng Hua}, {and} \bibinfo{person}{Lei Zhang}.} \bibinfo{year}{2022}\natexlab{}.
\newblock \showarticletitle{Box-supervised instance segmentation with level set evolution}. In \bibinfo{booktitle}{\emph{Proc. of ECCV}}.
\newblock


\bibitem[Li et~al\mbox{.}(2019)]%
        {Li_2019_ICCV}
\bibfield{author}{\bibinfo{person}{Xiaoyan Li}, \bibinfo{person}{Meina Kan}, \bibinfo{person}{Shiguang Shan}, {and} \bibinfo{person}{Xilin Chen}.} \bibinfo{year}{2019}\natexlab{}.
\newblock \showarticletitle{Weakly Supervised Object Detection With Segmentation Collaboration}. In \bibinfo{booktitle}{\emph{Proc. of ICCV}}.
\newblock


\bibitem[Li et~al\mbox{.}(2023)]%
        {zecheng2023cim}
\bibfield{author}{\bibinfo{person}{Zecheng Li}, \bibinfo{person}{Zening Zeng}, \bibinfo{person}{Yuqi Liang}, {and} \bibinfo{person}{Jin-Gang Yu}.} \bibinfo{year}{2023}\natexlab{}.
\newblock \showarticletitle{Complete Instances Mining for Weakly Supervised Instance Segmentation}. In \bibinfo{booktitle}{\emph{IJCAI}}.
\newblock


\bibitem[Liao et~al\mbox{.}(2019)]%
        {liao2019weakly}
\bibfield{author}{\bibinfo{person}{Shisha Liao}, \bibinfo{person}{Yongqing Sun}, \bibinfo{person}{Chenqiang Gao}, \bibinfo{person}{Pranav Shenoy~K. P}, \bibinfo{person}{Song Mu}, \bibinfo{person}{Jun Shimamura}, {and} \bibinfo{person}{Atsushi Sagata}.} \bibinfo{year}{2019}\natexlab{}.
\newblock \showarticletitle{Weakly Supervised Instance Segmentation Using Hybrid Networks}. In \bibinfo{booktitle}{\emph{{IEEE} International Conference on Acoustics, Speech and Signal Processing, {ICASSP} 2019, Brighton, United Kingdom, May 12-17, 2019}}.
\newblock


\bibitem[Lin et~al\mbox{.}(2024)]%
        {lin2023weakly}
\bibfield{author}{\bibinfo{person}{Jianghang Lin}, \bibinfo{person}{Yunhang Shen}, \bibinfo{person}{Bingquan Wang}, \bibinfo{person}{Shaohui Lin}, \bibinfo{person}{Ke Li}, {and} \bibinfo{person}{Liujuan Cao}.} \bibinfo{year}{2024}\natexlab{}.
\newblock \showarticletitle{Weakly Supervised Open-Vocabulary Object Detection}. In \bibinfo{booktitle}{\emph{AAAI}}.
\newblock


\bibitem[Lin et~al\mbox{.}(2017)]%
        {lin2017focal_loss}
\bibfield{author}{\bibinfo{person}{Tsung{-}Yi Lin}, \bibinfo{person}{Priya Goyal}, \bibinfo{person}{Ross~B. Girshick}, \bibinfo{person}{Kaiming He}, {and} \bibinfo{person}{Piotr Doll{\'{a}}r}.} \bibinfo{year}{2017}\natexlab{}.
\newblock \showarticletitle{Focal Loss for Dense Object Detection}. In \bibinfo{booktitle}{\emph{Proc. of ICCV}}.
\newblock


\bibitem[Lin et~al\mbox{.}(2014)]%
        {lin2014microsoft}
\bibfield{author}{\bibinfo{person}{Tsung-Yi Lin}, \bibinfo{person}{Michael Maire}, \bibinfo{person}{Serge Belongie}, \bibinfo{person}{James Hays}, \bibinfo{person}{Pietro Perona}, \bibinfo{person}{Deva Ramanan}, \bibinfo{person}{Piotr Doll{\'a}r}, {and} \bibinfo{person}{C~Lawrence Zitnick}.} \bibinfo{year}{2014}\natexlab{}.
\newblock \showarticletitle{Microsoft coco: Common objects in context}. In \bibinfo{booktitle}{\emph{Proc. of ECCV}}.
\newblock


\bibitem[Lin et~al\mbox{.}(2023)]%
        {lin2023clipes}
\bibfield{author}{\bibinfo{person}{Yuqi Lin}, \bibinfo{person}{Minghao Chen}, \bibinfo{person}{Wenxiao Wang}, \bibinfo{person}{Boxi Wu}, \bibinfo{person}{Ke Li}, \bibinfo{person}{Binbin Lin}, \bibinfo{person}{Haifeng Liu}, {and} \bibinfo{person}{Xiaofei He}.} \bibinfo{year}{2023}\natexlab{}.
\newblock \showarticletitle{Clip is also an efficient segmenter: A text-driven approach for weakly supervised semantic segmentation}. In \bibinfo{booktitle}{\emph{Proc. of CVPR}}.
\newblock


\bibitem[Liu et~al\mbox{.}(2019)]%
        {liu2019utilizing}
\bibfield{author}{\bibinfo{person}{Boxiao Liu}, \bibinfo{person}{Yan Gao}, \bibinfo{person}{Nan Guo}, \bibinfo{person}{Xiaochun Ye}, \bibinfo{person}{Fang Wan}, \bibinfo{person}{Haihang You}, {and} \bibinfo{person}{Dongrui Fan}.} \bibinfo{year}{2019}\natexlab{}.
\newblock \showarticletitle{Utilizing the Instability in Weakly Supervised Object Detection.}. In \bibinfo{booktitle}{\emph{Proc. of CVPR}}.
\newblock


\bibitem[Liu et~al\mbox{.}(2020)]%
        {liu2020liid}
\bibfield{author}{\bibinfo{person}{Yun Liu}, \bibinfo{person}{Yu-Huan Wu}, \bibinfo{person}{Peisong Wen}, \bibinfo{person}{Yujun Shi}, \bibinfo{person}{Yu Qiu}, {and} \bibinfo{person}{Ming-Ming Cheng}.} \bibinfo{year}{2020}\natexlab{}.
\newblock \showarticletitle{Leveraging instance-, image-and dataset-level information for weakly supervised instance segmentation}.
\newblock \bibinfo{journal}{\emph{IEEE TPAMI}} (\bibinfo{year}{2020}).
\newblock


\bibitem[Locatello et~al\mbox{.}(2020)]%
        {locatello2020wscomp}
\bibfield{author}{\bibinfo{person}{Francesco Locatello}, \bibinfo{person}{Ben Poole}, \bibinfo{person}{Gunnar R{\"{a}}tsch}, \bibinfo{person}{Bernhard Sch{\"{o}}lkopf}, \bibinfo{person}{Olivier Bachem}, {and} \bibinfo{person}{Michael Tschannen}.} \bibinfo{year}{2020}\natexlab{}.
\newblock \showarticletitle{Weakly-Supervised Disentanglement Without Compromises}. In \bibinfo{booktitle}{\emph{Proc. of ICML}}.
\newblock


\bibitem[Loshchilov and Hutter(2019)]%
        {loshchilov2017adamw}
\bibfield{author}{\bibinfo{person}{Ilya Loshchilov} {and} \bibinfo{person}{Frank Hutter}.} \bibinfo{year}{2019}\natexlab{}.
\newblock \showarticletitle{Decoupled Weight Decay Regularization}. In \bibinfo{booktitle}{\emph{Proc. of ICLR}}.
\newblock


\bibitem[Ma and Wang(2023)]%
        {ma2023medsam}
\bibfield{author}{\bibinfo{person}{Jun Ma} {and} \bibinfo{person}{Bo Wang}.} \bibinfo{year}{2023}\natexlab{}.
\newblock \showarticletitle{Segment anything in medical images}.
\newblock \bibinfo{journal}{\emph{ArXiv preprint}} (\bibinfo{year}{2023}).
\newblock


\bibitem[Ou et~al\mbox{.}(2021)]%
        {ou2021wsrcnn}
\bibfield{author}{\bibinfo{person}{Jia-Rong Ou}, \bibinfo{person}{Shu-Le Deng}, {and} \bibinfo{person}{Jin-Gang Yu}.} \bibinfo{year}{2021}\natexlab{}.
\newblock \showarticletitle{WS-RCNN: Learning to Score Proposals for Weakly Supervised Instance Segmentation}.
\newblock \bibinfo{journal}{\emph{Sensors}} (\bibinfo{year}{2021}).
\newblock


\bibitem[Pont-Tuset et~al\mbox{.}(2016)]%
        {pont2016mcg}
\bibfield{author}{\bibinfo{person}{Jordi Pont-Tuset}, \bibinfo{person}{Pablo Arbelaez}, \bibinfo{person}{Jonathan~T Barron}, \bibinfo{person}{Ferran Marques}, {and} \bibinfo{person}{Jitendra Malik}.} \bibinfo{year}{2016}\natexlab{}.
\newblock \showarticletitle{Multiscale combinatorial grouping for image segmentation and object proposal generation}.
\newblock \bibinfo{journal}{\emph{IEEE TPAMI}} (\bibinfo{year}{2016}).
\newblock


\bibitem[Qian and Zhang(2022)]%
        {mmQian2022wsss}
\bibfield{author}{\bibinfo{person}{Chen Qian} {and} \bibinfo{person}{Hui Zhang}.} \bibinfo{year}{2022}\natexlab{}.
\newblock \showarticletitle{Region-based Pixels Integration Mechanism for Weakly Supervised Semantic Segmentation}. In \bibinfo{booktitle}{\emph{ACM MM}}. \bibinfo{numpages}{9}~pages.
\newblock
\showISBNx{9781450392037}


\bibitem[Ren et~al\mbox{.}(2015)]%
        {ren2015faster}
\bibfield{author}{\bibinfo{person}{Shaoqing Ren}, \bibinfo{person}{Kaiming He}, \bibinfo{person}{Ross~B. Girshick}, {and} \bibinfo{person}{Jian Sun}.} \bibinfo{year}{2015}\natexlab{}.
\newblock \showarticletitle{Faster {R-CNN:} Towards Real-Time Object Detection with Region Proposal Networks}. In \bibinfo{booktitle}{\emph{Proc. of NeurIPS}}.
\newblock


\bibitem[Ren et~al\mbox{.}(2020)]%
        {ren2020instance}
\bibfield{author}{\bibinfo{person}{Zhongzheng Ren}, \bibinfo{person}{Zhiding Yu}, \bibinfo{person}{Xiaodong Yang}, \bibinfo{person}{Ming{-}Yu Liu}, \bibinfo{person}{Yong~Jae Lee}, \bibinfo{person}{Alexander~G. Schwing}, {and} \bibinfo{person}{Jan Kautz}.} \bibinfo{year}{2020}\natexlab{}.
\newblock \showarticletitle{Instance-Aware, Context-Focused, and Memory-Efficient Weakly Supervised Object Detection}. In \bibinfo{booktitle}{\emph{Proc. of CVPR}}.
\newblock


\bibitem[Schroeter et~al\mbox{.}(2019)]%
        {schroeter2019wstemp}
\bibfield{author}{\bibinfo{person}{Julien Schroeter}, \bibinfo{person}{Kirill~A. Sidorov}, {and} \bibinfo{person}{A.~David Marshall}.} \bibinfo{year}{2019}\natexlab{}.
\newblock \showarticletitle{Weakly-Supervised Temporal Localization via Occurrence Count Learning}. In \bibinfo{booktitle}{\emph{Proc. of ICML}}.
\newblock


\bibitem[Seo et~al\mbox{.}(2022)]%
        {seo2022object}
\bibfield{author}{\bibinfo{person}{Jinhwan Seo}, \bibinfo{person}{Wonho Bae}, \bibinfo{person}{Danica~J Sutherland}, \bibinfo{person}{Junhyug Noh}, {and} \bibinfo{person}{Daijin Kim}.} \bibinfo{year}{2022}\natexlab{}.
\newblock \showarticletitle{Object discovery via contrastive learning for weakly supervised object detection}. In \bibinfo{booktitle}{\emph{Proc. of ECCV}}.
\newblock


\bibitem[Shao et~al\mbox{.}(2021)]%
        {mmShao2021wsol}
\bibfield{author}{\bibinfo{person}{Feifei Shao}, \bibinfo{person}{Yawei Luo}, \bibinfo{person}{Li Zhang}, \bibinfo{person}{Lu Ye}, \bibinfo{person}{Siliang Tang}, \bibinfo{person}{Yi Yang}, {and} \bibinfo{person}{Jun Xiao}.} \bibinfo{year}{2021}\natexlab{}.
\newblock \showarticletitle{Improving Weakly Supervised Object Localization via Causal Intervention}. In \bibinfo{booktitle}{\emph{ACM MM}}. \bibinfo{numpages}{9}~pages.
\newblock
\showISBNx{9781450386517}


\bibitem[Shen et~al\mbox{.}(2021)]%
        {shen2021parallel}
\bibfield{author}{\bibinfo{person}{Yunhang Shen}, \bibinfo{person}{Liujuan Cao}, \bibinfo{person}{Zhiwei Chen}, \bibinfo{person}{Baochang Zhang}, \bibinfo{person}{Chi Su}, \bibinfo{person}{Yongjian Wu}, \bibinfo{person}{Feiyue Huang}, {and} \bibinfo{person}{Rongrong Ji}.} \bibinfo{year}{2021}\natexlab{}.
\newblock \showarticletitle{Parallel Detection-and-Segmentation Learning for Weakly Supervised Instance Segmentation}. In \bibinfo{booktitle}{\emph{Proc. of ICCV}}.
\newblock


\bibitem[Shen et~al\mbox{.}(2020)]%
        {shen2020enabling}
\bibfield{author}{\bibinfo{person}{Yunhang Shen}, \bibinfo{person}{Rongrong Ji}, \bibinfo{person}{Yan Wang}, \bibinfo{person}{Zhiwei Chen}, \bibinfo{person}{Feng Zheng}, \bibinfo{person}{Feiyue Huang}, {and} \bibinfo{person}{Yunsheng Wu}.} \bibinfo{year}{2020}\natexlab{}.
\newblock \showarticletitle{Enabling deep residual networks for weakly supervised object detection}. In \bibinfo{booktitle}{\emph{Proc. of ECCV}}.
\newblock


\bibitem[Shen et~al\mbox{.}(2019)]%
        {shen2019cyclic}
\bibfield{author}{\bibinfo{person}{Yunhang Shen}, \bibinfo{person}{Rongrong Ji}, \bibinfo{person}{Yan Wang}, \bibinfo{person}{Yongjian Wu}, {and} \bibinfo{person}{Liujuan Cao}.} \bibinfo{year}{2019}\natexlab{}.
\newblock \showarticletitle{Cyclic Guidance for Weakly Supervised Joint Detection and Segmentation}. In \bibinfo{booktitle}{\emph{Proc. of CVPR}}.
\newblock


\bibitem[Sui et~al\mbox{.}(2022)]%
        {sui2022sos-wsods}
\bibfield{author}{\bibinfo{person}{Lin Sui}, \bibinfo{person}{Chen-Lin Zhang}, {and} \bibinfo{person}{Jianxin Wu}.} \bibinfo{year}{2022}\natexlab{}.
\newblock \showarticletitle{Salvage of supervision in weakly supervised object detection}. In \bibinfo{booktitle}{\emph{Proc. of CVPR}}.
\newblock


\bibitem[Sun et~al\mbox{.}(2020)]%
        {sun2020mining}
\bibfield{author}{\bibinfo{person}{Guolei Sun}, \bibinfo{person}{Wenguan Wang}, \bibinfo{person}{Jifeng Dai}, {and} \bibinfo{person}{Luc Van~Gool}.} \bibinfo{year}{2020}\natexlab{}.
\newblock \showarticletitle{Mining cross-image semantics for weakly supervised semantic segmentation}. In \bibinfo{booktitle}{\emph{Proc. of ECCV}}.
\newblock


\bibitem[Sun et~al\mbox{.}(2023)]%
        {sun2023samwsss}
\bibfield{author}{\bibinfo{person}{Weixuan Sun}, \bibinfo{person}{Zheyuan Liu}, \bibinfo{person}{Yanhao Zhang}, \bibinfo{person}{Yiran Zhong}, {and} \bibinfo{person}{Nick Barnes}.} \bibinfo{year}{2023}\natexlab{}.
\newblock \showarticletitle{An Alternative to WSSS? An Empirical Study of the Segment Anything Model (SAM) on Weakly-Supervised Semantic Segmentation Problems}.
\newblock \bibinfo{journal}{\emph{ArXiv preprint}} (\bibinfo{year}{2023}).
\newblock


\bibitem[Tan et~al\mbox{.}(2020)]%
        {mmTan2020wsol}
\bibfield{author}{\bibinfo{person}{Chuangchuang Tan}, \bibinfo{person}{Guanghua Gu}, \bibinfo{person}{Tao Ruan}, \bibinfo{person}{Shikui Wei}, {and} \bibinfo{person}{Yao Zhao}.} \bibinfo{year}{2020}\natexlab{}.
\newblock \showarticletitle{Dual-Gradients Localization Framework for Weakly Supervised Object Localization}. In \bibinfo{booktitle}{\emph{{MM} '20: The 28th {ACM} International Conference on Multimedia, Virtual Event / Seattle, WA, USA, October 12-16, 2020}}.
\newblock


\bibitem[Tang et~al\mbox{.}(2018a)]%
        {tang2018pcl}
\bibfield{author}{\bibinfo{person}{Peng Tang}, \bibinfo{person}{Xinggang Wang}, \bibinfo{person}{Song Bai}, \bibinfo{person}{Wei Shen}, \bibinfo{person}{Xiang Bai}, \bibinfo{person}{Wenyu Liu}, {and} \bibinfo{person}{Alan Yuille}.} \bibinfo{year}{2018}\natexlab{a}.
\newblock \showarticletitle{Pcl: Proposal cluster learning for weakly supervised object detection}.
\newblock \bibinfo{journal}{\emph{IEEE TPAMI}} (\bibinfo{year}{2018}).
\newblock


\bibitem[Tang et~al\mbox{.}(2017)]%
        {tang2017multiple}
\bibfield{author}{\bibinfo{person}{Peng Tang}, \bibinfo{person}{Xinggang Wang}, \bibinfo{person}{Xiang Bai}, {and} \bibinfo{person}{Wenyu Liu}.} \bibinfo{year}{2017}\natexlab{}.
\newblock \showarticletitle{Multiple Instance Detection Network with Online Instance Classifier Refinement}. In \bibinfo{booktitle}{\emph{Proc. of CVPR}}.
\newblock


\bibitem[Tang et~al\mbox{.}(2018b)]%
        {tang2018weakly}
\bibfield{author}{\bibinfo{person}{Peng Tang}, \bibinfo{person}{Xinggang Wang}, \bibinfo{person}{Angtian Wang}, \bibinfo{person}{Yongluan Yan}, \bibinfo{person}{Wenyu Liu}, \bibinfo{person}{Junzhou Huang}, {and} \bibinfo{person}{Alan Yuille}.} \bibinfo{year}{2018}\natexlab{b}.
\newblock \showarticletitle{Weakly supervised region proposal network and object detection}. In \bibinfo{booktitle}{\emph{Proc. of ECCV}}.
\newblock


\bibitem[Tian et~al\mbox{.}(2021)]%
        {tian2021boxinst}
\bibfield{author}{\bibinfo{person}{Zhi Tian}, \bibinfo{person}{Chunhua Shen}, \bibinfo{person}{Xinlong Wang}, {and} \bibinfo{person}{Hao Chen}.} \bibinfo{year}{2021}\natexlab{}.
\newblock \showarticletitle{BoxInst: High-Performance Instance Segmentation With Box Annotations}. In \bibinfo{booktitle}{\emph{Proc. of CVPR}}.
\newblock


\bibitem[Touvron et~al\mbox{.}(2021)]%
        {touvron2021deit}
\bibfield{author}{\bibinfo{person}{Hugo Touvron}, \bibinfo{person}{Matthieu Cord}, \bibinfo{person}{Matthijs Douze}, \bibinfo{person}{Francisco Massa}, \bibinfo{person}{Alexandre Sablayrolles}, {and} \bibinfo{person}{Herv{\'{e}} J{\'{e}}gou}.} \bibinfo{year}{2021}\natexlab{}.
\newblock \showarticletitle{Training data-efficient image transformers {\&} distillation through attention}. In \bibinfo{booktitle}{\emph{Proc. of ICML}}.
\newblock


\bibitem[Uijlings et~al\mbox{.}(2013)]%
        {Uijlings2013SelectiveSF}
\bibfield{author}{\bibinfo{person}{Jasper R.~R. Uijlings}, \bibinfo{person}{Koen E.~A. van~de Sande}, \bibinfo{person}{Theo Gevers}, {and} \bibinfo{person}{Arnold W.~M. Smeulders}.} \bibinfo{year}{2013}\natexlab{}.
\newblock \showarticletitle{Selective Search for Object Recognition}.
\newblock \bibinfo{journal}{\emph{IJCV}} (\bibinfo{year}{2013}).
\newblock


\bibitem[Wan et~al\mbox{.}(2019)]%
        {wan2019c}
\bibfield{author}{\bibinfo{person}{Fang Wan}, \bibinfo{person}{Chang Liu}, \bibinfo{person}{Wei Ke}, \bibinfo{person}{Xiangyang Ji}, \bibinfo{person}{Jianbin Jiao}, {and} \bibinfo{person}{Qixiang Ye}.} \bibinfo{year}{2019}\natexlab{}.
\newblock \showarticletitle{{C-MIL:} Continuation Multiple Instance Learning for Weakly Supervised Object Detection}. In \bibinfo{booktitle}{\emph{Proc. of CVPR}}.
\newblock


\bibitem[Wan et~al\mbox{.}(2018)]%
        {wan2018min}
\bibfield{author}{\bibinfo{person}{Fang Wan}, \bibinfo{person}{Pengxu Wei}, \bibinfo{person}{Jianbin Jiao}, \bibinfo{person}{Zhenjun Han}, {and} \bibinfo{person}{Qixiang Ye}.} \bibinfo{year}{2018}\natexlab{}.
\newblock \showarticletitle{Min-Entropy Latent Model for Weakly Supervised Object Detection}. In \bibinfo{booktitle}{\emph{Proc. of CVPR}}.
\newblock


\bibitem[Wang et~al\mbox{.}(2021)]%
        {wang2021boxcaseg}
\bibfield{author}{\bibinfo{person}{Xinggang Wang}, \bibinfo{person}{Jiapei Feng}, \bibinfo{person}{Bin Hu}, \bibinfo{person}{Qi Ding}, \bibinfo{person}{Longjin Ran}, \bibinfo{person}{Xiaoxin Chen}, {and} \bibinfo{person}{Wenyu Liu}.} \bibinfo{year}{2021}\natexlab{}.
\newblock \showarticletitle{Weakly-Supervised Instance Segmentation via Class-Agnostic Learning With Salient Images}. In \bibinfo{booktitle}{\emph{Proc. of CVPR}}.
\newblock


\bibitem[Wang et~al\mbox{.}(2013)]%
        {wang2013mmmil}
\bibfield{author}{\bibinfo{person}{Xinggang Wang}, \bibinfo{person}{Baoyuan Wang}, \bibinfo{person}{Xiang Bai}, \bibinfo{person}{Wenyu Liu}, {and} \bibinfo{person}{Zhuowen Tu}.} \bibinfo{year}{2013}\natexlab{}.
\newblock \showarticletitle{Max-Margin Multiple-Instance Dictionary Learning}. In \bibinfo{booktitle}{\emph{Proc. of ICML}}.
\newblock


\bibitem[Xu et~al\mbox{.}(2014)]%
        {xu2014lmws}
\bibfield{author}{\bibinfo{person}{Chang Xu}, \bibinfo{person}{Dacheng Tao}, \bibinfo{person}{Chao Xu}, {and} \bibinfo{person}{Yong Rui}.} \bibinfo{year}{2014}\natexlab{}.
\newblock \showarticletitle{Large-margin Weakly Supervised Dimensionality Reduction}. In \bibinfo{booktitle}{\emph{Proc. of ICML}}.
\newblock


\bibitem[Xu et~al\mbox{.}(2022b)]%
        {mmXu2022wsol}
\bibfield{author}{\bibinfo{person}{Jingyuan Xu}, \bibinfo{person}{Hongtao Xie}, \bibinfo{person}{Chuanbin Liu}, {and} \bibinfo{person}{Yongdong Zhang}.} \bibinfo{year}{2022}\natexlab{b}.
\newblock \showarticletitle{Proxy Probing Decoder for Weakly Supervised Object Localization: A Baseline Investigation}. In \bibinfo{booktitle}{\emph{ACM MM}}.
\newblock
\showISBNx{9781450392037}


\bibitem[Xu et~al\mbox{.}(2022c)]%
        {mmXu2022wsss}
\bibfield{author}{\bibinfo{person}{Jianjun Xu}, \bibinfo{person}{Hongtao Xie}, \bibinfo{person}{Hai Xu}, \bibinfo{person}{Yuxin Wang}, \bibinfo{person}{Sun-ao Liu}, {and} \bibinfo{person}{Yongdong Zhang}.} \bibinfo{year}{2022}\natexlab{c}.
\newblock \showarticletitle{Boat in the Sky: Background Decoupling and Object-aware Pooling for Weakly Supervised Semantic Segmentation}. In \bibinfo{booktitle}{\emph{ACM MM}}.
\newblock
\showISBNx{9781450392037}


\bibitem[Xu et~al\mbox{.}(2022a)]%
        {xu2022mctformer}
\bibfield{author}{\bibinfo{person}{Lian Xu}, \bibinfo{person}{Wanli Ouyang}, \bibinfo{person}{Mohammed Bennamoun}, \bibinfo{person}{Farid Boussaid}, {and} \bibinfo{person}{Dan Xu}.} \bibinfo{year}{2022}\natexlab{a}.
\newblock \showarticletitle{Multi-class token transformer for weakly supervised semantic segmentation}. In \bibinfo{booktitle}{\emph{Proc. of CVPR}}.
\newblock


\bibitem[Yan et~al\mbox{.}(2019)]%
        {gao2019c}
\bibfield{author}{\bibinfo{person}{Gao Yan}, \bibinfo{person}{Boxiao Liu}, \bibinfo{person}{Nan Guo}, \bibinfo{person}{Xiaochun Ye}, \bibinfo{person}{Fang Wan}, \bibinfo{person}{Haihang You}, {and} \bibinfo{person}{Dongrui Fan}.} \bibinfo{year}{2019}\natexlab{}.
\newblock \showarticletitle{{C-MIDN:} Coupled Multiple Instance Detection Network With Segmentation Guidance for Weakly Supervised Object Detection}. In \bibinfo{booktitle}{\emph{Proc. of ICCV}}.
\newblock


\bibitem[Yang et~al\mbox{.}(2019)]%
        {yang2019towards}
\bibfield{author}{\bibinfo{person}{Ke Yang}, \bibinfo{person}{Dongsheng Li}, {and} \bibinfo{person}{Yong Dou}.} \bibinfo{year}{2019}\natexlab{}.
\newblock \showarticletitle{Towards Precise End-to-End Weakly Supervised Object Detection Network}. In \bibinfo{booktitle}{\emph{Proc. of ICCV}}.
\newblock


\bibitem[Yang et~al\mbox{.}(2020)]%
        {mmYang2020wsod}
\bibfield{author}{\bibinfo{person}{Ke Yang}, \bibinfo{person}{Peng Zhang}, \bibinfo{person}{Peng Qiao}, \bibinfo{person}{Zhiyuan Wang}, \bibinfo{person}{Dongsheng Li}, {and} \bibinfo{person}{Yong Dou}.} \bibinfo{year}{2020}\natexlab{}.
\newblock \showarticletitle{Objectness Consistent Representation for Weakly Supervised Object Detection}. In \bibinfo{booktitle}{\emph{{MM} '20: The 28th {ACM} International Conference on Multimedia, Virtual Event / Seattle, WA, USA, October 12-16, 2020}}.
\newblock


\bibitem[Yang et~al\mbox{.}(2023)]%
        {yang2023fgvp}
\bibfield{author}{\bibinfo{person}{Lingfeng Yang}, \bibinfo{person}{Yueze Wang}, \bibinfo{person}{Xiang Li}, \bibinfo{person}{Xinlong Wang}, {and} \bibinfo{person}{Jian Yang}.} \bibinfo{year}{2023}\natexlab{}.
\newblock \showarticletitle{Fine-Grained Visual Prompting}.
\newblock \bibinfo{journal}{\emph{ArXiv preprint}} (\bibinfo{year}{2023}).
\newblock


\bibitem[Yin et~al\mbox{.}(2021)]%
        {yin2021instance}
\bibfield{author}{\bibinfo{person}{Yufei Yin}, \bibinfo{person}{Jiajun Deng}, \bibinfo{person}{Wengang Zhou}, {and} \bibinfo{person}{Houqiang Li}.} \bibinfo{year}{2021}\natexlab{}.
\newblock \showarticletitle{Instance Mining with Class Feature Banks for Weakly Supervised Object Detection}. In \bibinfo{booktitle}{\emph{Proc. of AAAI}}.
\newblock


\bibitem[Yin et~al\mbox{.}(2023)]%
        {yin2023cyclic}
\bibfield{author}{\bibinfo{person}{Yufei Yin}, \bibinfo{person}{Jiajun Deng}, \bibinfo{person}{Wengang Zhou}, \bibinfo{person}{Li Li}, {and} \bibinfo{person}{Houqiang Li}.} \bibinfo{year}{2023}\natexlab{}.
\newblock \showarticletitle{Cyclic-Bootstrap Labeling for Weakly Supervised Object Detection}. In \bibinfo{booktitle}{\emph{ICCV}}.
\newblock


\bibitem[Zeng et~al\mbox{.}(2019)]%
        {zeng2019wsod2}
\bibfield{author}{\bibinfo{person}{Zhaoyang Zeng}, \bibinfo{person}{Bei Liu}, \bibinfo{person}{Jianlong Fu}, \bibinfo{person}{Hongyang Chao}, {and} \bibinfo{person}{Lei Zhang}.} \bibinfo{year}{2019}\natexlab{}.
\newblock \showarticletitle{{WSOD2:} Learning Bottom-Up and Top-Down Objectness Distillation for Weakly-Supervised Object Detection}. In \bibinfo{booktitle}{\emph{Proc. of ICCV}}.
\newblock


\bibitem[Zhang et~al\mbox{.}(2022b)]%
        {zhang2022dino}
\bibfield{author}{\bibinfo{person}{Hao Zhang}, \bibinfo{person}{Feng Li}, \bibinfo{person}{Shilong Liu}, \bibinfo{person}{Lei Zhang}, \bibinfo{person}{Hang Su}, \bibinfo{person}{Jun Zhu}, \bibinfo{person}{Lionel~M. Ni}, {and} \bibinfo{person}{Heung-Yeung Shum}.} \bibinfo{year}{2022}\natexlab{b}.
\newblock \bibinfo{title}{DINO: DETR with Improved DeNoising Anchor Boxes for End-to-End Object Detection}.
\newblock
\newblock
\showeprint[arxiv]{2203.03605}~[cs.CV]


\bibitem[Zhang et~al\mbox{.}(2023)]%
        {zhang2023weakly}
\bibfield{author}{\bibinfo{person}{Jiabin Zhang}, \bibinfo{person}{Hu Su}, \bibinfo{person}{Yonghao He}, {and} \bibinfo{person}{Wei Zou}.} \bibinfo{year}{2023}\natexlab{}.
\newblock \showarticletitle{Weakly Supervised Instance Segmentation via Category-aware Centerness Learning with Localization Supervision}.
\newblock \bibinfo{journal}{\emph{PR}} (\bibinfo{year}{2023}).
\newblock


\bibitem[Zhang et~al\mbox{.}(2021b)]%
        {zhang2021weakly}
\bibfield{author}{\bibinfo{person}{Ke Zhang}, \bibinfo{person}{Chun Yuan}, \bibinfo{person}{Yiming Zhu}, \bibinfo{person}{Yong Jiang}, {and} \bibinfo{person}{Lishu Luo}.} \bibinfo{year}{2021}\natexlab{b}.
\newblock \showarticletitle{Weakly supervised instance segmentation by exploring entire object regions}.
\newblock \bibinfo{journal}{\emph{IEEE TMM}} (\bibinfo{year}{2021}).
\newblock


\bibitem[Zhang et~al\mbox{.}(2022a)]%
        {mmZhang2022wsss}
\bibfield{author}{\bibinfo{person}{Meijie Zhang}, \bibinfo{person}{Jianwu Li}, {and} \bibinfo{person}{Tianfei Zhou}.} \bibinfo{year}{2022}\natexlab{a}.
\newblock \showarticletitle{Multi-Granular Semantic Mining for Weakly Supervised Semantic Segmentation}. In \bibinfo{booktitle}{\emph{ACM MM}}.
\newblock
\showISBNx{9781450392037}


\bibitem[Zhang et~al\mbox{.}(2018b)]%
        {zhang2018zigzag}
\bibfield{author}{\bibinfo{person}{Xiaopeng Zhang}, \bibinfo{person}{Jiashi Feng}, \bibinfo{person}{Hongkai Xiong}, {and} \bibinfo{person}{Qi Tian}.} \bibinfo{year}{2018}\natexlab{b}.
\newblock \showarticletitle{Zigzag Learning for Weakly Supervised Object Detection}. In \bibinfo{booktitle}{\emph{Proc. of CVPR}}.
\newblock


\bibitem[Zhang et~al\mbox{.}(2021a)]%
        {mmZhang2021wsss}
\bibfield{author}{\bibinfo{person}{Xiangrong Zhang}, \bibinfo{person}{Zelin Peng}, \bibinfo{person}{Peng Zhu}, \bibinfo{person}{Tianyang Zhang}, \bibinfo{person}{Chen Li}, \bibinfo{person}{Huiyu Zhou}, {and} \bibinfo{person}{Licheng Jiao}.} \bibinfo{year}{2021}\natexlab{a}.
\newblock \showarticletitle{Adaptive Affinity Loss and Erroneous Pseudo-Label Refinement for Weakly Supervised Semantic Segmentation}. In \bibinfo{booktitle}{\emph{ACM MM}}.
\newblock
\showISBNx{9781450386517}


\bibitem[Zhang et~al\mbox{.}(2018a)]%
        {zhang2018w2f}
\bibfield{author}{\bibinfo{person}{Yongqiang Zhang}, \bibinfo{person}{Yancheng Bai}, \bibinfo{person}{Mingli Ding}, \bibinfo{person}{Yongqiang Li}, {and} \bibinfo{person}{Bernard Ghanem}.} \bibinfo{year}{2018}\natexlab{a}.
\newblock \showarticletitle{{W2F:} {A} Weakly-Supervised to Fully-Supervised Framework for Object Detection}. In \bibinfo{booktitle}{\emph{Proc. of CVPR}}.
\newblock


\bibitem[Zhou et~al\mbox{.}(2016)]%
        {zhou2015learning}
\bibfield{author}{\bibinfo{person}{Bolei Zhou}, \bibinfo{person}{Aditya Khosla}, \bibinfo{person}{{\`{A}}gata Lapedriza}, \bibinfo{person}{Aude Oliva}, {and} \bibinfo{person}{Antonio Torralba}.} \bibinfo{year}{2016}\natexlab{}.
\newblock \showarticletitle{Learning Deep Features for Discriminative Localization}. In \bibinfo{booktitle}{\emph{Proc. of CVPR}}.
\newblock


\bibitem[Zhou et~al\mbox{.}(2018)]%
        {zhou2018prm}
\bibfield{author}{\bibinfo{person}{Yanzhao Zhou}, \bibinfo{person}{Yi Zhu}, \bibinfo{person}{Qixiang Ye}, \bibinfo{person}{Qiang Qiu}, {and} \bibinfo{person}{Jianbin Jiao}.} \bibinfo{year}{2018}\natexlab{}.
\newblock \showarticletitle{Weakly Supervised Instance Segmentation Using Class Peak Response}. In \bibinfo{booktitle}{\emph{Proc. of CVPR}}.
\newblock


\bibitem[Zhou(2018)]%
        {zhou2018wsl_intro}
\bibfield{author}{\bibinfo{person}{Zhi-Hua Zhou}.} \bibinfo{year}{2018}\natexlab{}.
\newblock \showarticletitle{A brief introduction to weakly supervised learning}.
\newblock \bibinfo{journal}{\emph{National science review}} (\bibinfo{year}{2018}).
\newblock


\bibitem[Zhu et~al\mbox{.}(2023a)]%
        {zhu2023weaktr}
\bibfield{author}{\bibinfo{person}{Lianghui Zhu}, \bibinfo{person}{Yingyue Li}, \bibinfo{person}{Jieming Fang}, \bibinfo{person}{Yan Liu}, \bibinfo{person}{Hao Xin}, \bibinfo{person}{Wenyu Liu}, {and} \bibinfo{person}{Xinggang Wang}.} \bibinfo{year}{2023}\natexlab{a}.
\newblock \showarticletitle{WeakTr: Exploring Plain Vision Transformer for Weakly-supervised Semantic Segmentation}.
\newblock \bibinfo{journal}{\emph{ArXiv preprint}} (\bibinfo{year}{2023}).
\newblock


\bibitem[Zhu et~al\mbox{.}(2023b)]%
        {zhu2023encoder}
\bibfield{author}{\bibinfo{person}{Liangjun Zhu}, \bibinfo{person}{Li Peng}, \bibinfo{person}{Shuchen Ding}, {and} \bibinfo{person}{Zhongren Liu}.} \bibinfo{year}{2023}\natexlab{b}.
\newblock \showarticletitle{An encoder-decoder framework with dynamic convolution for weakly supervised instance segmentation}.
\newblock \bibinfo{journal}{\emph{IET Computer Vision}} (\bibinfo{year}{2023}).
\newblock


\bibitem[Zhu et~al\mbox{.}(2019)]%
        {Zhu2019iam}
\bibfield{author}{\bibinfo{person}{Yi Zhu}, \bibinfo{person}{Yanzhao Zhou}, \bibinfo{person}{Huijuan Xu}, \bibinfo{person}{Qixiang Ye}, \bibinfo{person}{David~S. Doermann}, {and} \bibinfo{person}{Jianbin Jiao}.} \bibinfo{year}{2019}\natexlab{}.
\newblock \showarticletitle{Learning Instance Activation Maps for Weakly Supervised Instance Segmentation}. In \bibinfo{booktitle}{\emph{Proc. of CVPR}}.
\newblock


\bibitem[Zitnick and Doll{\'a}r(2014)]%
        {zitnick2014eb}
\bibfield{author}{\bibinfo{person}{C~Lawrence Zitnick} {and} \bibinfo{person}{Piotr Doll{\'a}r}.} \bibinfo{year}{2014}\natexlab{}.
\newblock \showarticletitle{Edge boxes: Locating object proposals from edges}. In \bibinfo{booktitle}{\emph{Proc. of ECCV}}.
\newblock


\end{thebibliography}

\newpage
\appendix

\section{More Details of \name{}~Proposals}
\label{sec: md_wks_proposals}

\begin{figure}[ht]
  \centering
  \setlength{\tabcolsep}{1.6 pt}
  \includegraphics[width=1\linewidth]{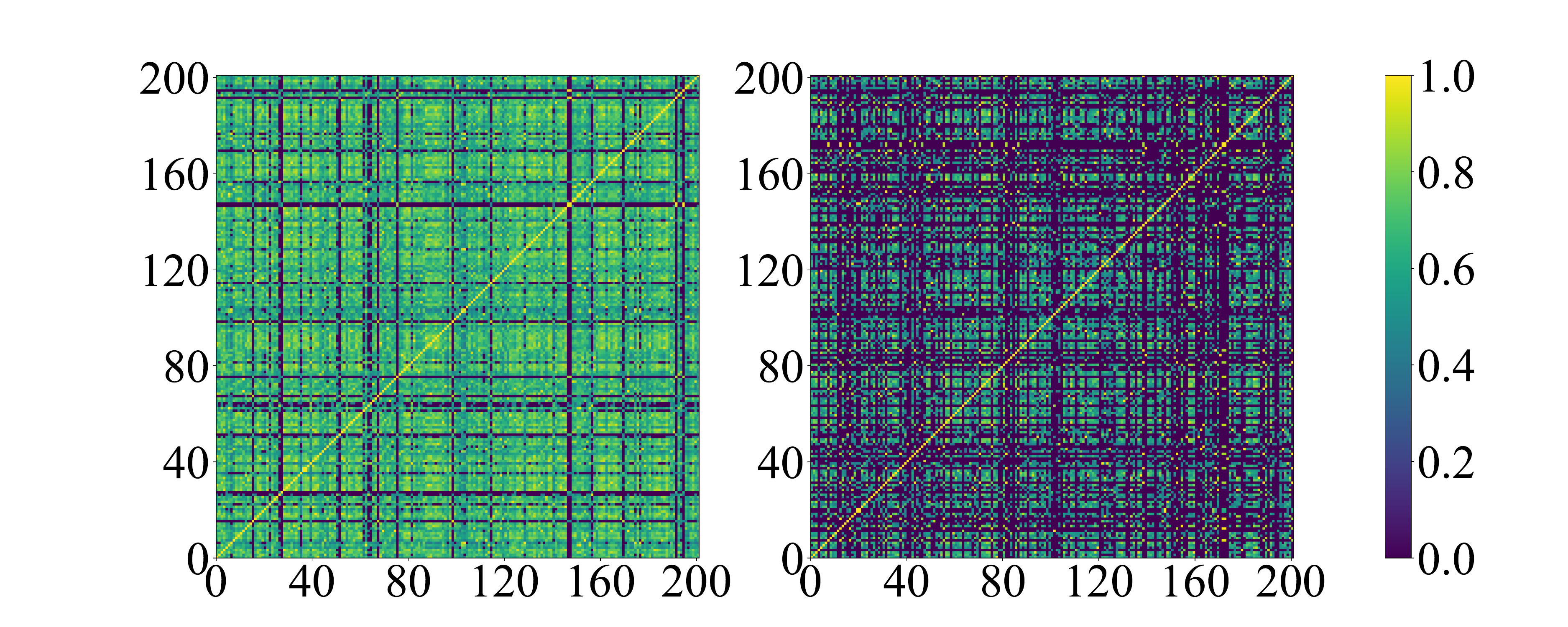}
  \caption{
  The cosine similarity among the features of proposals, i.e., Left for Selective Search proposals and Right for \name{} proposals. For a single image from PASCAL VOC 2007, we randomly sampled 200 proposal features to calculate their similarity.}
   \label{fig:comp_proposals}
   \vspace{-0.1 in}
\end{figure}

We further analyze the proposal similarity in the different weakly-supervised object detection (WSOD) proposals, as shown in Fig.~\ref{fig:comp_proposals}. We randomly sample 200 proposal features each from Selective Search~\cite{Uijlings2013SelectiveSF} proposals and \name{} proposals, and then compute their cosine similarity, respectively. Please note that all the features are output by the RoI pooling layer. It can be seen that the features from Selective Search tend to have higher similarity with other ones. In contrast, the features from \name{} proposals show lower similarity, which usually means it has less overlap and redundancy.

\section{More Details of RoI Drop Regularization}
\label{sec: md_roi_drop}

\begin{figure}[ht]
  \vspace{-0.1 in}
  \centering
  \setlength{\tabcolsep}{1.6 pt}
  \includegraphics[width=1\linewidth]{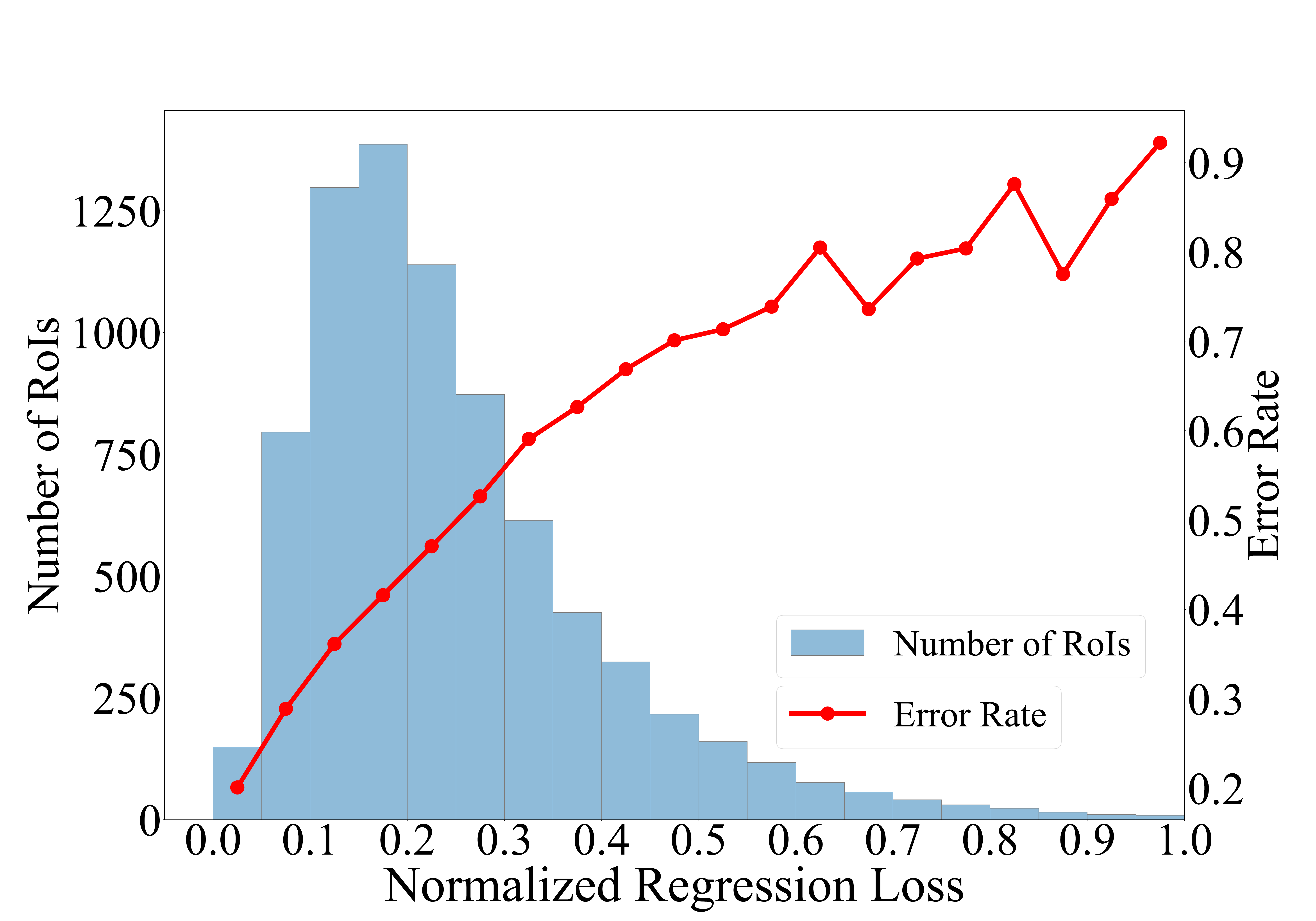}
  \caption{
  The relationship between the normalized regression loss, the corresponding number of RoIs, and the corresponding error rate. The results are obtained from training the Faster-RCNN~\cite{ren2015faster} using PASCAL VOC 2007 pseudo ground truth (PGT) in the preliminary training stage.}
   \label{fig:loss_drop_reg}
   \vspace{-0.1 in}
\end{figure}

We further present the relationship between the normalized regression loss, the corresponding number of RoIs, and the corresponding error rate in Fig.~\ref{fig:loss_drop_reg}. It shows that the regression losses of RoIs have different number distributions compared to the classification losses. However, they exhibit similar error rate curves. This observation further demonstrates the necessity of RoI drop regularization with a regression threshold $\tau_{reg}$.

\section{More Details of Query Drop Regularization}
\label{sec: md_query_drop}

Because DINO~\cite{zhang2022dino} employs Focal loss~\cite{lin2017focal_loss} as the classification loss, queries associated with background classes tend to have higher predicted probabilities and lower losses. This results in the inadvertent omission of most foreground category queries when directly dropping queries. To mitigate this issue, our first step involves normalizing the unweighted Focal loss, which is essentially the binary cross-entropy loss, for both foreground and background queries within each training batch. Normalizing at the batch level broadens the sampling scope from a single image to the size of the batch. In the second step, queries are dropped based on their loss ranking post-normalization. This approach avoids making the model converge slowly due to the dropping of the most foreground queries.

\subsection{Ablation Studies for RoI Drop Regularization}
\begin{table}[htp]
    \small
    \renewcommand\arraystretch{1.2}
    \caption{Ablation study for the regression threshold and classification threshold in RoI drop regularization in terms of $\text{AP}_{\scriptsize50}$ on the PASCAL VOC 2007 $test$ set.}
    \centering

    \begin{subtable}{.45\linewidth} %
    \caption{Ablation study for the regression threshold.}
    \centering
    \begin{tabular}{c|ccc}
    \Xhline{1pt}
     $\tau_{reg}$ & 0.8 & 1.0 & 1.2 \\
    \Xhline{0.5pt}
    $\text{AP}_{\scriptsize50}$ & 71.0 & 71.8 & 71.3 \\
    \Xhline{1pt}
    \end{tabular}
    
    \end{subtable}%
    \hfill
    \begin{subtable}{.45\linewidth} %
    \caption{Ablation study for the classification threshold.}
    \centering
    \begin{tabular}{c|ccc}
    \Xhline{1pt}
    $\tau_{cls}$ & 3.0 & 4.0 & 5.0 \\
    \Xhline{0.5pt}
    $\text{AP}_{\scriptsize50}$ & 71.2 & 71.8 & 71.1 \\
    \Xhline{1pt}
    \end{tabular}
    \end{subtable}

    \label{tab: abla_roi_drop}
\end{table}

To further analyze the impact of the regression threshold and classification threshold in RoI drop regularization, we conduct experiments as shown in Table~\ref{tab: abla_roi_drop}. It is observed that the best regression threshold $\tau_{reg}$ and classification threshold $\tau_{cls}$ for RoI drop regularization is 1.0 and 4.0, respectively.

\subsection{Ablation Studies for Query Drop Regularization}

\begin{table}[htp]
\small
\setlength{\tabcolsep}{6pt}
\caption{Ablation studies for query drop regularization on the PASCAL VOC 2007 $test$ set.}
\centering
\begin{tabular}{ccc|c}
\toprule
Baseline & $\text{Query}_\text{things}$ & $\text{Query}_\text{bkg}$ & \text{AP}$_{50}$ \\
\midrule
 \Checkmark       &&& 72.8 \\
\midrule
         &\Checkmark       &         &     73.4\scriptsize\color{red}+0.6         \\
         &\Checkmark       &    \Checkmark     &  73.3\scriptsize\color{red}+0.5  \\

\bottomrule
\end{tabular}
\label{tab: abla_query_fore_or_bkg}
\end{table}

\begin{figure*}[ht]

    \centering
    \includegraphics[width=1.0\linewidth]{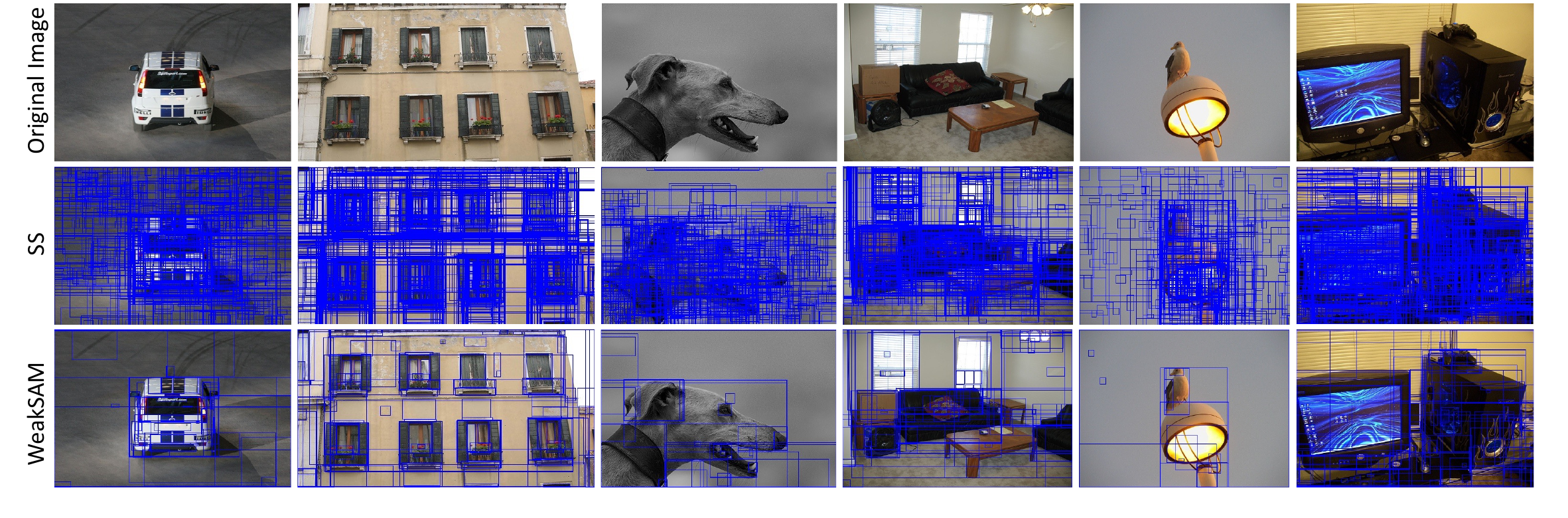}
    \vspace{-.3 in}
    \caption{Visualization of the proposals boxes on the PASCAL VOC 2007 $trainval$ set.}
    \label{fig: vis_proposals}
\vspace{-0.2cm}
\end{figure*}

\begin{figure*}[ht]

    \centering
    \includegraphics[width=1.0\linewidth]{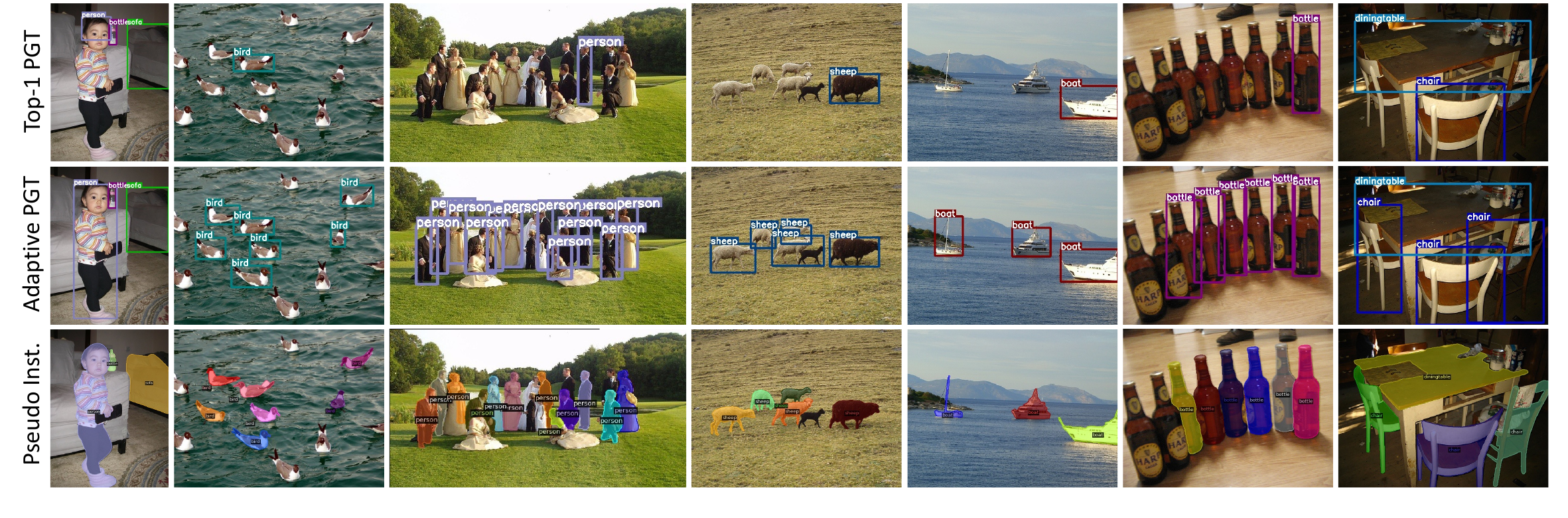}
    \vspace{-.3 in}
    \caption{Visualization of the pseudo ground truth boxes and pseudo instance labels on the PASCAL VOC 2012 $trainaug$ set.}
    \label{fig: vis_pgt}
\end{figure*}

To analyze the impact of dropping queries corresponding to foreground categories $\text{Query}_\text{things}$ and background categories $\text{Query}_\text{bkg}$, we conduct ablations as shown in Table~\ref{tab: abla_query_fore_or_bkg}. Experimental results indicate that dropping only the foreground queries ($\text{Query}_\text{things}$) leads to the best performance, whereas dropping both types of queries results in a slight decrease in performance. We maintain the viewpoint that dropping more background queries may also lead to slower convergence. Consequently, we choose to drop only $\text{Query}_\text{things}$ to achieve better performance.

We further analyze the impact of classification threshold $\tau$ in query drop regularization, as shown in Table~\ref{tab: abla_query_drop}. Quantitative results demonstrate that 90 is the best percentile classification threshold.

\begin{table}[htp]
    \renewcommand\arraystretch{1.}
    \caption{Ablation study for the classification threshold in query drop regularization in terms of $\text{AP}_{\scriptsize50}$ on the PASCAL VOC 2007 $test$ set.}
    \centering
    \begin{tabular}{c|ccc}
    \Xhline{1pt}
     $\tau$ (\%) & 100 & 90 & 80 \\
    \Xhline{0.5pt}
    $\text{AP}_{\scriptsize50}$ & 72.8 & 73.4 & 71.8 \\
    \Xhline{1pt}
    \end{tabular}

    \label{tab: abla_query_drop}
\end{table}

\section{Additional Quantitative Results}
\label{sec: add_quan}

\begin{table}[htp]
\small
\centering
\renewcommand\arraystretch{1}
\caption{Comparison on PASCAL VOC 2007 trainval set in terms of CorLoc(\%) with multi-scale testing.}
\begin{tabular}{lccc}
\toprule
\textbf{Methods}&                        $Sup.$& \textbf{Proposal}& \textbf{CorLoc}\\
\midrule
WSDDN~\cite{bilen2016wsddn}&   \multirow{10}{*}{$\mathcal{I}$}                   & EB~\cite{zitnick2014eb}& 53.5\\ %
Yang et al.~\cite{yang2019towards} &                 & SS~\cite{Uijlings2013SelectiveSF} & 68.0\\ %
C-MIL ~\cite{wan2019c}&                   & SS& 65.0\\ %
C-MIDN~\cite{gao2019c}&                     & SS& 53.5\\ %
WSOD2~\cite{zeng2019wsod2}&                      & SS& 69.5\\ %
CASD~\cite{huang2020comprehensive}&                       & SS& 70.4\\ %
OD-WSCL~\cite{seo2022object}&                   & SS& 69.8\\ %
WSOD-CBL~\cite{yin2023cyclic}&                & SS& 71.8\\ %
WSOVOD~\cite{lin2023weakly} &   & LO-WSRPN+SAM & 77.2 \\
WSOVOD\ddag &  & LO-WSRPN+SAM & 80.1 \\
\midrule
OICR~\cite{tang2017multiple}&    \multirow{2}{*}{$\mathcal{I}$}    & SS& 60.6\\ %
\name{} (OICR)&                             & \name{}& 74.5\scriptsize\color{red}+13.9\\
\midrule
MIST~\cite{ren2020instance}&     \multirow{2}{*}{$\mathcal{I}$}   & SS& 68.8\\ %
\name{} (MIST)&                             & \name{}& 82.9\scriptsize\color{red}+14.1\\
\toprule
\end{tabular}
\label{tab:corloc}
\end{table}

We present the comparison on PASCAL VOC 2007 $trainval$ set in terms of CorLoc, as shown in Table~\ref{tab:corloc}. It can be seen that the \name{} achieves the 13.9\% and 14.1\% CorLoc improvements on OICR and MIST, respectively. The \name{} (OICR) outperforms the WSOD-CBL~\cite{yin2023cyclic} by 11.1\% CorLoc. The results demonstrate the significant performance improvement brought by our \name{}.

\section{Additional Ablation Studies}

\label{sec: add_abla}

\subsection{Analysis on Classification Methods}
\begin{table}[ht]
\centering
\setlength{\tabcolsep}{4pt}
\caption{Ablation studies for classification methods that generate \name{} queries on PASCAL VOC 2007 $trainval$ set. We evaluate the average number of proposals and Recall. We present the results of Selective Search~\cite{Uijlings2013SelectiveSF} at the first row as a baseline.}
\vspace{0.1in}
\begin{tabular}{l|c|ccc}
\toprule
  \multirow{2}{*}{CLS Methods} & \multirow{2}{*}{Num.} & \multicolumn{3}{c}{Recall} \\
& & IoU=0.50 & IoU=0.75 & IoU=0.90 \\
\midrule
None & 2001 & 92.6 & 57.7 & 19.2 \\
\midrule
WeakTr~\cite{zhu2023weaktr} &   213   &    95.6   &   75.0    &   42.1 \\ %
MCTformer~\cite{xu2022mctformer} &   173   &    93.2   &    74.8   &   43.7  \\ %
CLIP-ES~\cite{lin2023clipes} &   205   &    93.8   &  75.6     &    44.5 \\ %

\bottomrule
\end{tabular}
\vspace{-0.1in}
\label{tab: abla_wsss}
\end{table}

To further analyze the impact of methods that generate classification clues, we replaced WeakTr in \name{} with MCTformer and CLIP-ES. As indicated in Table~\ref{tab: abla_wsss}, \name{} (MCTformer) achieves a 1.6\% higher Recall (IoU=90) than \name{} (WeakTr). Furthermore, \name{} (CLIP-ES) records increases of 0.6\% and 2.4\% in Recall over \name{} (WeakTr) at IoU thresholds of 75 and 90, respectively. These results demonstrate the versatility of the \name{} proposal-generating method across different classification methods. Please note that all classification methods employed in this study are CAM networks from weakly-supervised semantic segmentation (WSSS) methods. Since these networks are typically well-tuned on specific datasets, such as PASCAL VOC 2012 and COCO 2014, they are adept at providing rich classification clues.
\vspace{-.1in}

\section{Additional Visualization Results}
\label{sec: add_vis}

Fig.\ref{fig: vis_proposals} compares the Selective Search proposals with those generated by \name{}. The \name{} proposals exhibit less redundancy than Selective Search proposals. Fig.\ref{fig: vis_pgt} contrasts the Top-1 PGT with adaptive PGT, demonstrating that adaptive PGT generation captures a greater number of target objects, which might be missed by the Top-1 approach. Additionally, adaptive PGT can be seamlessly integrated to generate pseudo instance labels.

\section{More Implementation Details}
\label{sec: add_impl}

For Algorithm. 1 in paper, 
we set the kernel size $k$ to 128 and activation threshold $\tau$ to 0.9 following default parameters from WeakTr~\cite{zhu2023weaktr}. 
And for Algorithm. 2, 
we follow the default manners similar to SoS-WSOD~\cite{sui2022sos-wsods}, in which score threshold $\tau_s$ is set to 0.3, and overlap threshold $\tau_o$ is set to 0.85.

For Faster R-CNN~\cite{ren2015faster} retraining, we adopt the same training strategy and hyper-parameters as the fully-supervised ones.
For DINO~\cite{zhang2022dino} retraining, we use a learning rate of 9e-5 with the AdamW~\cite{loshchilov2017adamw} optimizer, and a max epoch of 14. Moreover, we apply multi-scale augmentation and horizontal flips in both training and testing.

For implementations of Mask R-CNN~\cite{he2017mask} and Mask2Former~\cite{cheng2022mask2f}, we follow their default hyper-parameters.

\end{document}